\documentclass[10pt,twocolumn,letterpaper]{article}

\usepackage{cvpr}              
\usepackage[accsupp]{axessibility} 

\usepackage{graphicx}
\usepackage{amsmath}
\usepackage{amssymb}
\usepackage{booktabs}
\usepackage{algorithm}
\usepackage{algorithmic}
\usepackage{multirow}
\usepackage{multicol}
\usepackage{diagbox}
\usepackage{amsmath,amssymb}

\usepackage[pagebackref,breaklinks,colorlinks]{hyperref}

\usepackage[capitalize]{cleveref}
\crefname{section}{Sec.}{Secs.}
\Crefname{section}{Section}{Sections}
\Crefname{table}{Table}{Tables}
\crefname{table}{Tab.}{Tabs.}

\def\cvprPaperID{3039} 
\def\confName{CVPR}
\def\confYear{2023}

\begin{document}

\title{Explicit Boundary Guided Semi-Push-Pull Contrastive Learning for Supervised Anomaly Detection}

\author{
Xincheng Yao$^{{\rm 1}}$, Ruoqi Li$^{{\rm 1}}$, Jing Zhang$^{{\rm 3}}$, Jun Sun$^{{\rm 1}}$, Chongyang Zhang$^{{\rm 1},{\rm 2}}\thanks{Corresponding Author.}$\\
\textsuperscript{\rm 1}School of Electronic Information and Electrical Engineering, Shanghai Jiao Tong University\\
\textsuperscript{\rm 2}MoE Key Lab of Artificial Intelligence, AI Institute, Shanghai Jiao Tong University\\
\textsuperscript{\rm 3}Research Institute of Systems Engineering, Academy Military Science, Beijing, China\\
{\tt\small \{i-Dover, nilponi, junsun, sunny\_zhang\}@sjtu.edu.cn$^{{\rm 1}}$}, {\tt\small jingzhang@163.com.cn$^{{\rm 3}}$}
}

\maketitle

\begin{abstract}
    Most anomaly detection (AD) models are learned using only normal samples in an unsupervised way, which may result in ambiguous decision boundary and insufficient discriminability. In fact, a few anomaly samples are often available in real-world applications, the valuable knowledge of known anomalies should also be effectively exploited. However, utilizing a few known anomalies during training may cause another issue that the model may be biased by those known anomalies and fail to generalize to unseen anomalies.
   In this paper, we tackle supervised anomaly detection, \emph{i.e.}, we learn AD models using a few available anomalies with the objective to detect both the seen and unseen anomalies. We propose a novel explicit boundary guided semi-push-pull contrastive learning mechanism, which can enhance model's discriminability while mitigating the bias issue. Our approach is based on two core designs: First, we find an explicit and compact separating boundary as the guidance for further feature learning. As the boundary only relies on the normal feature distribution, the bias problem caused by a few known anomalies can be alleviated. Second, a boundary guided semi-push-pull loss is developed to only pull the normal features together while pushing the abnormal features apart from the separating boundary beyond a certain margin region. In this way, our model can form a more explicit and discriminative decision boundary to distinguish known and also unseen anomalies from normal samples more effectively.
   Code will be available at \url{https://github.com/xcyao00/BGAD}.
\end{abstract}

\section{Introduction}
\label{sec:intro}
 
\begin{figure*}[ht]
     \centering
     \includegraphics[width=1.0\linewidth]{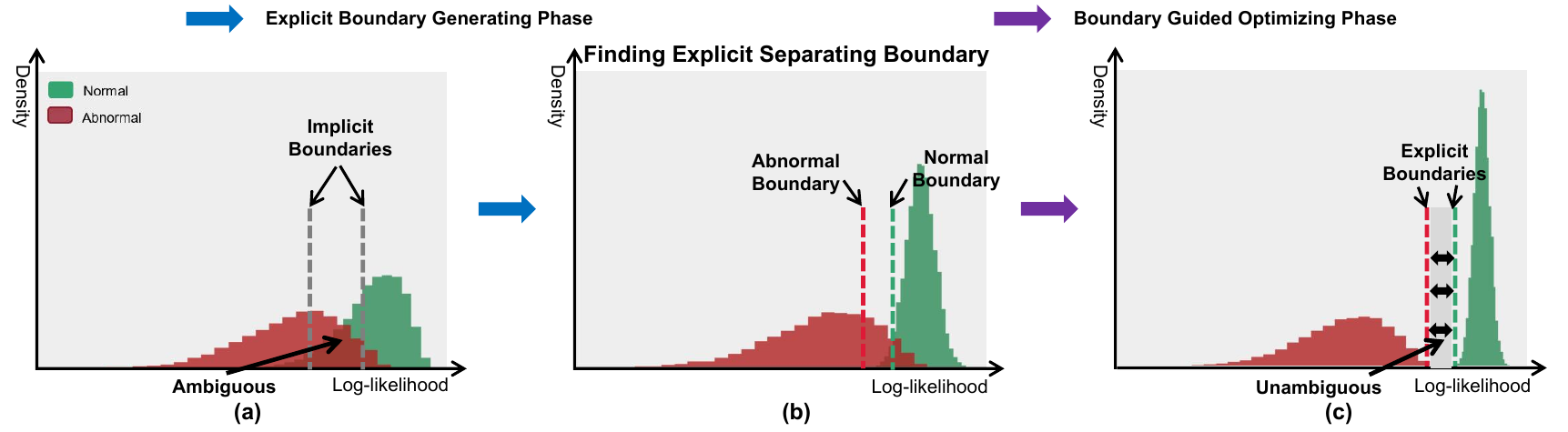} 
    \caption{Conceptual illustration of our method. (a) In most unsupervised AD models, the anomaly score distribution usually has ambiguous regions, which makes it difficult to get one ideal decision boundary. \emph{E.g.}, the left boundary will cause many false negatives, while the right boundary may induce many false positives. (b) With the normalized normal feature distribution, a pair of explicit and compact (close to the normal distribution) boundaries can be obtained easily. (c) With the proposed BG-SPP loss, boundary guided optimizing can be implemented to obtain an unambiguous anomaly score distribution: a significant gap between the normal and abnormal distributions.}
    \label{fig:activation}
 \end{figure*}

 Anomaly detection (AD) has received widespread attention in diverse domains, such as industrial defect inspection \cite{MVTec, PaDiM, PatchSVDD, SPADE} and medical lesion detection \cite{KDAD, DRA}. Most previous anomaly detection methods \cite{GANomaly, SSIM, PatchSVDD, DeepKNN, SPADE, PaDiM, MSFD, CFLOW, FastFlow, DRAEM, STAD, DifferNet, PANDA} are unsupervised and pay much attention to normal samples while inadvertently overlooking anomalies, because it is difficult to collect sufficient and all kinds of anomalies. However, learning only from normal samples may limit the discriminability of the AD models \cite{FCDD, DRA}. As illustrated in Figure \ref{fig:activation}(a), without anomalies, the decision boundaries are generally implicit and not discriminative enough. The \emph{insufficient discriminability} issue is a common issue in unsupervised anomaly detection due to the lack of knowledge about anomalies. In fact, a few anomalies are usually available in real-world applications, which can be exploited effectively to address or alleviate this issue.

 
Recently, methods that can be called semi-supervised AD \cite{SAD, HSC, DevNet} or AD with outlier exposure \cite{OE, OE2} begin to focus on those available anomalies. These methods attempt to learn knowledge from anomalies by one-class classification with anomalies as negative samples \cite{SAD, HSC} or by supervised binary classification \cite{OE, OE2} or by utilizing the deviation loss to optimize one anomaly scoring network \cite{DevNet}. They show a fact that the detection performance can be improved significantly even with a few anomalies. However, the known anomalies can't represent all kinds of anomalies. These methods may be biased by the known anomalies and fail to generalize to unseen anomalies (see Figure \ref{fig:feature_tsne}). 

Therefore, to address the two above issues, we tackle supervised anomaly detection \cite{DRA}, in which a few known anomalies can be effectively exploited to train discriminative AD models with the objective to improve detection performance on the known anomalies and generalize well to unseen anomalies. Compared with unsupervised AD, supervised AD is more meaningful for real-world AD applications, because the detected anomalies can be used to further improve the discriminability and generalizability of the model. To this end, we propose a novel \textbf{B}oundary \textbf{G}uided \textbf{A}nomaly \textbf{D}etection (\textbf{BGAD}) model, which has two core designs as illustrated in Figure \ref{fig:activation}: explicit boundary generating and boundary guided optimizing. 

$\bullet$ \textbf{Explicit Boundary Generating.} We first employ normalizing flow \cite{realNVP} to learn a normalized normal feature distribution, and obtain an explicit separating boundary, which is close to the normal feature distribution edge and controlled by a hyperparameter $\beta$ (\emph{i.e.}, the normal boundary in Figure \ref{fig:activation}(b)). The obtained explicit separating boundary only relies on the normal distribution and has no relation with the abnormal samples, thus the bias problem caused by the insufficient known anomalies can be mitigated. 

$\bullet$ \textbf{Boundary Guided Optimizing.} After obtaining the explicit separating boundary, we then propose a boundary guided semi-push-pull (BG-SPP) loss to exploit anomalies for learning more discriminative features. With the BG-SPP loss, only the normal features whose log-likelihoods are smaller than the boundary are pulled together to form a more compact normal feature distribution (semi-pull); while the abnormal features whose log-likelihoods are larger than the boundary are pushed apart from the boundary beyond a certain margin region (semi-push). 

In this way, our model can form a more explicit and discriminative separating boundary and also a reliable margin region for distinguishing anomalies more effectively (see Figure \ref{fig:activation}(c), \ref{fig:histgram}).  Furthermore, rarity is a critical problem of anomalies and may cause feature learning inefficient. We thus propose RandAugment-based Pseudo Anomaly Generation, which can simulate anomalies by creating local irregularities in normal samples, to tackle the rarity challenge. 

In summary, we make the following main contributions:

1. We propose a novel Explicit Boundary Guided supervised AD modeling method, in which both normal and abnormal samples are exploited effectively by well-designed explicit boundary generating and boundary guided optimizing. With the proposed AD method, higher discriminability and lower bias risk can be achieved simultaneously.

2. To exploit a few known anomalies effectively, we propose a BG-SPP loss to pull together normal features while pushing abnormal features apart from the separating boundary, thus more discriminative features can be learned.

3. We achieve SOTA results on the widely-used MVTecAD benchmark, with the performance of 99.3\% image-level AUROC and 99.2\% pixel-level AUROC. 

\section{Related Work}
\label{sec:related_work}

\textbf{Unsupervised Approaches.} Most anomaly detection methods are unsupervised and only learn from normal samples, such as AutoEncoder \cite{SSIM, AutoEncoder2, PMAD}, GAN \cite{AnoGAN, fast-AnoGAN, GANomaly, ALAD, GAN1, GAN2} and one-class-classification (OCC) based methods \cite{OneclassSVM, SVDD, deepSVDD, PatchSVDD}. Recently, most superior methods utilize pre-trained deep models, such as DeepKNN \cite{DeepKNN}, GaussianAD \cite{PaDiM1}, SPADE \cite{SPADE}, PaDiM \cite{PaDiM} and PatchCore \cite{PatchCore}. There are also some anomaly detection methods based on knowledge distillation \cite{STAD, MSFD, KDAD}, feature reconstruction \cite{DFR}, and normalizing flows \cite{ResidualFlow, DifferNet, CFLOW, FastFlow}. Our method is significantly different from these works \cite{ResidualFlow, DifferNet, CFLOW, FastFlow} in the following three aspects: (1) \textbf{New Motivation}: our work aims
to learn knowledge from a few anomalies to address the insufficient discriminability issue and also mitigate the bias
issue. (2) \textbf{Novel Method}: we only use the normalizing flow model as a basic likelihood estimation
network, and are the first to propose Explicit Separating
Boundary and BG-SPP loss to achieve higher discriminability and also lower bias.  (3) \textbf{Different Task}: \cite{ResidualFlow} focuses on out-of-distribution detection and \cite{DifferNet, CFLOW, FastFlow} focus on unsupervised anomaly detection (AD), whereas our work focuses on supervised AD.

\textbf{Supervised Approaches.} Currently, a few existing works are similar to ours, \emph{i.e.}, AD with outlier exposure \cite{OE, OE2} and deep semi-supervised AD \cite{SAD, HSC, FCDD}. In \cite{OE}, Hendrycks, \emph{et al.} term random nature images from the large scale datasets that are likely not nominal as outlier exposure, and explore how to utilize such data to improve unsupervised AD. The method presented in \cite{OE2} utilizes thousands of OE samples to achieve state-of-the-art results on standard image AD benchmarks. DeepSAD \cite{SAD} is the first deep model utilizing a few anomalies by generalizing the unsupervised DeepSVDD \cite{deepSVDD} method to a semi-supervised AD setting. In \cite{HSC}, Ruff, \emph{et al.} further modify the DeepSAD based on cross-entropy classification that concentrates nominal samples, this modification significantly improves the performance of DeepSAD. FCDD proposed in \cite{FCDD} extends the pseudo-Huber loss in \cite{HSC} to construct a semi-supervised anomaly localization framework. Some works\cite{DevNet, DevNet2} utilize the deviation loss to optimize an anomaly scoring network, in which the anomaly scores of normal samples are imposed to approximate scalar scores drawn from the prior while that of anomaly samples are enforced to have significant deviations from these normal scores. However, these methods simply push abnormal features apart from the normal patterns as far as possible, which may cause bias in the model for the known anomalies. The recent work DRA \cite{DRA} is the most similar to ours, which also considers the model's generalization to unseen anomalies. The DRA model can learn disentangled representations of anomalies to enable generalizable detection. 



\section{Our Proposed Approach}
\label{sec:overview}

\begin{figure*}[ht]
    \centering
    \includegraphics[width=1.0\linewidth]{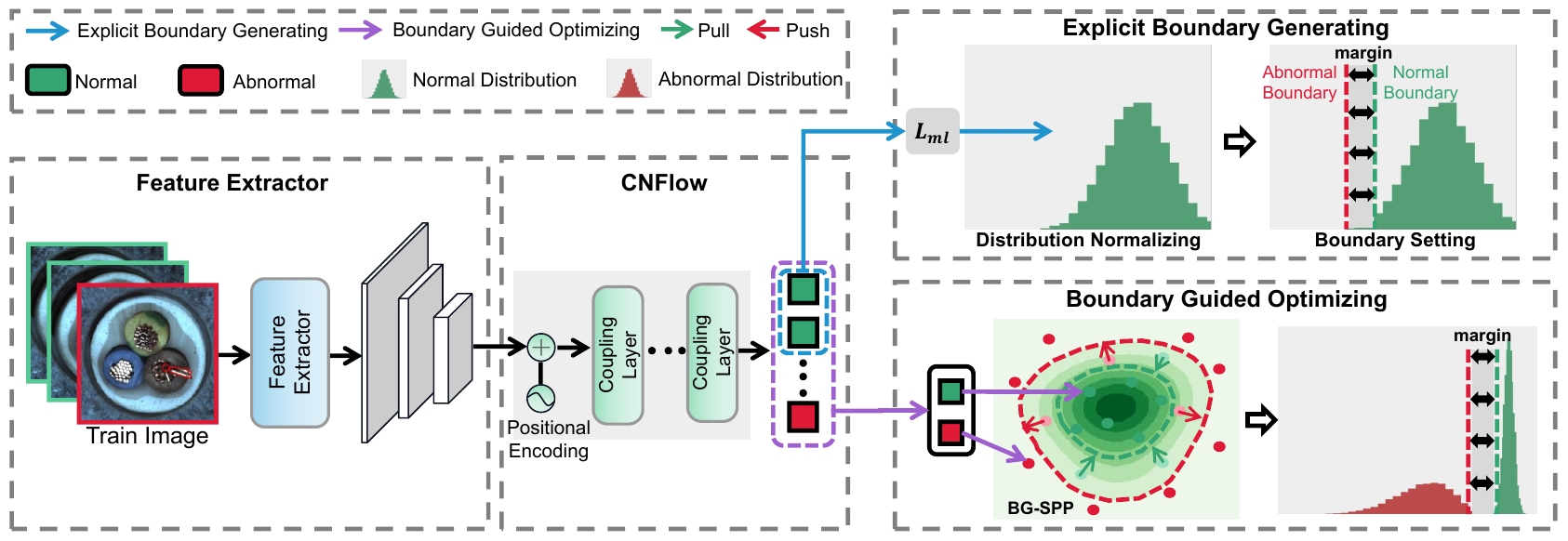} \\
    \caption{Model overview. The extracted feature maps are transformed into a latent space using a conditional normalizing flow (CNFlow), which is then used to generate an anomaly score for each feature. The training procedure can be divided into two phases: explicit boundary generating and boundary guided optimizing. In the first phase, only normal samples and ML loss (E.q.(\ref{eq:ml_loss})) are utilized for model training to obtain a relatively stable normal log-likelihood distribution, and then one explicit separating boundary can be obtained based on the learned distribution. In the second phase, with the explicit separating boundary and the BG-SPP loss (E.q.(\ref{eq:bg_sppc})), both normal and abnormal samples are utilized for model training to learn more discriminative features.}
    \label{fig:framework}
\end{figure*}

\textbf{Problem Statement.} Different from the general unsupervised AD setting, the training set of supervised AD is composed of normal images and a few anomalies, denoted as $\mathcal{I}_{train}=\mathcal{I}^n \cup \mathcal{I}^a$, where $\mathcal{I}^n = \{I^n_i\}_{i=1}^{N_0}$ and $\mathcal{I}^a = \{I^a_j\}_{j=1}^{M_0} (M_0 \ll N_0)$ indicate the collection of normal samples and abnormal samples. The $M_0$ anomalies are randomly sampled from the seen anomaly classes $\mathcal{S}_s \subset \mathcal{S}$, where $\mathcal{S} = \mathcal{S}_s \cup \mathcal{S}_u$ means all the seen and unseen anomaly classes. The goal is to learn a model $m : \mathcal{I} \rightarrow \mathbb{R}$ that can assign larger anomaly scores to both seen and unseen anomalies than normal samples.

\textbf{Overview.} Figure \ref{fig:framework} overviews our proposed method. The model consists of four parts: Feature Extractor $f: \mathcal{I} \rightarrow \mathcal{X}$, Conditional Normalizing Flow (CNFlow) $\varphi_\theta: \mathcal{X} \rightarrow \mathcal{Z}$, Explicit Boundary Generating and Boundary Guided Optimizing. We refer the features extracted by the feature extractor as input features for CNFlow, and denote these features as $\mathcal{X} = \mathcal{X}^n \cup \mathcal{X}^a$, where $\mathcal{X}^n=\{x^n_i\}_{i=1}^N$ and $\mathcal{X}^a=\{x^a_j\}_{j=1}^M$ are normal and abnormal features, respectively. The training procedure can be divided into two phases as shown in Figure \ref{fig:framework}: explicit boundary generating and boundary guided optimizing. 
In the testing procedure, the CNFlow can assign corresponding log-likelihoods for input features, and the log-likelihoods can be converted to anomaly scores (see Sec. \ref{sec:boundary_selection}). 

\label{sec:method}
\subsection{Learning Normal Feature Distribution by Normalizing Flow}
\label{sec:distribution_normalizing}
 In order to find one anomaly-independent separating boundary,
one simplified distribution of normal features should
be learned firstly. Normalizing flow \cite{NICE, realNVP} is employed to learn normal feature distribution in our method. 

\textbf{Conditional Normalizing Flow.} Formally, we denote $\varphi_\theta: \mathcal{X} \in \mathbb{R}^d \rightarrow \mathcal{Z} \in \mathbb{R}^d$ as our normalizing flow. It is built as a composition of coupling layers \cite{NICE} such that $\varphi_\theta=\varphi_L\circ\dots\circ\varphi_2\circ\varphi_1$, where $\theta$ is the trainable parameters and $L$ is the total number of layers. Defining $d$-dimensional input and output features of normalizing flow as $y_0=x\in \mathcal{X}$ and $y_L=z \in \mathcal{Z}$, the latents can be computed as $y_l=\varphi_l(y_{l-1})$, where $\{y_l\}_{l=1}^{L-1}$ are the intermediate outputs. The input distribution estimated by model $p_\theta(x)$ can be calculated according to the change of variables formula as follows \cite{NICE, realNVP}:
\begin{equation}
\label{eq:log_likelihood1}
    {\rm log}p_{\theta}(x)={\rm log}p_\mathcal{Z}(\varphi_\theta(x))+\sum\nolimits_{l=1}^{L}{\rm log}\big|{\rm det}J_{\varphi_l}(y_{l-1})\big|
\end{equation}
where $J_{\varphi_l}(y_{l-1}) = \frac{\partial \varphi_l(y_{l-1})}{\partial y_{l-1}}$ is the Jacobian matrix of the transformation $\varphi_l$ at $y_{l-1}$, and ${\rm det}$ means determinant. Normalizing flow can approximate the feature distribution $p_{\mathcal{X}}$ with $p_\theta(x)$. The set of parameters $\theta$ is obtained by optimizing the log-likelihoods across the training distribution $p_{\mathcal{X}}$:
\begin{equation}
\label{eq:maximum_optimization}
    \theta^* = \mathop{{\rm argmin}}\limits_{\theta \in \Theta}\mathbb{E}_{x \sim p_{\mathcal{X}}}[-{\rm log}p_\theta(x)]
\end{equation}

 The coupling layers in normalizing flow are usually implemented by fully connected layers, so the spatial position relationship will be destroyed because the 2D feature maps are flattened to 1D. To preserve the positional information, we follow \cite{CFLOW} to add 2D-aware position embeddings. 

\textbf{Learning Normal Feature Distribution.} We then employ normalizing flow to learn normal feature distribution by maximum likelihood optimization. The latent variable distribution $p_\mathcal{Z}(z), z \in \mathbb{R}^d$ can generally be assumed to obey the multivariate Gaussian distribution \cite{CFLOW} as follows:
\begin{equation}
\label{eq:gaussian}
    p_\mathcal{Z}(z) = (2\pi)^{-\frac{d}{2}}{\rm det}(\Sigma^{-\frac{1}{2}}){\rm e}^{-\frac{1}{2}(z-\mu)^T\Sigma^{-1}(z-\mu)}
\end{equation}
 where $\mu$ is the mean and $\Sigma$ is the covariance. When training normal features, the latent variables for normal features can be assumed to obey $\mathcal{N}(0, \mathbf{I})$ for further simplicity. By replacing $p_\mathcal{Z}(z)=(2\pi)^{-\frac{d}{2}}{\rm e}^{-\frac{1}{2}z^Tz}$ in formula (\ref{eq:log_likelihood1}), the optimization objective in the formula (\ref{eq:maximum_optimization}) can be rewritten as:
\begin{align}
    \theta^* = \mathop{{\rm argmin}}\limits_{\theta \in \Theta}\mathbb{E}_{x \sim p_\mathcal{X}}\Big[&\frac{d}{2}{\rm log}(2\pi)+\frac{1}{2}\varphi_\theta(x)^T\varphi_\theta(x) \nonumber \\
    &-\sum\nolimits_{l=1}^{L}{\rm log}\big|{\rm det}J_{\varphi_l}(y_{l-1})|\Big]
\end{align}

The maximum likelihood loss function for learning normal feature distribution can be defined as:
\begin{align}
    \label{eq:ml_loss}
    \mathcal{L}_{ml}= \mathbb{E}_{x \in \mathcal{X}^n}\Big[\frac{d}{2}{\rm log}&(2\pi)+\frac{1}{2}\varphi_\theta(x)^T\varphi_\theta(x) \nonumber \\
    &-\sum\nolimits_{l=1}^{L}{\rm log}\big|{\rm det}J_{\varphi_l}(y_{l-1})|\Big]
\end{align}

\subsection{Finding an Explicit and Compact Separating Boundary}
\label{sec:boundary_selection}

With the learned normal feature distribution, we can further find one explicit and compact separating boundary.  However, due to the high dimensional characteristics of the features, we therefore consider finding the boundary from the anomaly score distribution. Since the loglikelihoods
generated by the CNFlow can be equivalently
converted to anomaly scores, we select the boundary on the
log-likelihood distribution.

\textbf{Anomaly Scoring.} The advantage of normalizing flow is that we can estimate the exact log-likelihood ${\rm log}p(x)$ for each input feature $x$ as follows:
\begin{align}
\label{eq:log_likelihood2}
    {\rm log}p(x) = -\frac{d}{2}{\rm log}(2\pi)&-\frac{1}{2}\varphi_\theta(x)^T\varphi_\theta(x) \nonumber \\
    &+\sum\nolimits_{l=1}^{L}{\rm log}\big|{\rm det}J_{\varphi_l}(y_{l-1})|
\end{align}

 With the estimated log-likelihood ${\rm log}p(x)$, we can convert it to likelihood via exponential function. As we maximize log-likelihoods for normal features in E.q.(\ref{eq:ml_loss}), the likelihood can directly measure the normality. Thus, we can generate the anomaly score as follows:
\begin{equation}
    s(x) = 1 - {\rm exp}({\rm log}p(x))
\end{equation}
where the $s(x)$ means the anomaly score of $x$. Because exponential function is monotonic, the log-likelihood can be equivalently converted to the anomaly score. Thus, the separating boundary in log-likelihood distribution is equivalent to the boundary in anomaly score distribution.  

\textbf{Finding Explicit Separating Boundary.} We then obtain the separating boundary based on log-likelihood distribution. We build the boundary through the following steps:

\textbf{1. Building normal log-likelihood distribution.} We can employ the log-likelihood estimation formulation in E.q.(\ref{eq:log_likelihood2}) to obtain all log-likelihoods of normal features $\mathcal{P}_n=\{{\rm log}p_i\}_{i=1}^N$. The $\mathcal{P}_n$ can be used to approximate the log-likelihood distribution of all normal features. 

\textbf{2. Finding explicit normal and abnormal boundary.} How to find a suitable boundary is a dilemma. If we set the boundary too close to the distribution center, the samples in the tail of the normal distribution are more likely to be misclassified as abnormal. Meanwhile, if the boundary is far away from the distribution center, more anomalies would be determined as normal ones. Thus, we define a position hyperparameter $\beta$ to control the distance from the boundary to the center. We select the $\beta$-th percentile (\emph{e.g.}, $\beta=1$) of sorted normal log-likelihood distribution as the normal boundary $b_n$, which also indicates the upper bound of the normal false positive rate is $\beta\%$. To make the feature learning more robust, we further introduce a margin hyperparameter $\tau$ (\emph{e.g.}, $\tau=0.1$) and define an abnormal boundary $b_a = b_n - \tau$ (see Figure \ref{fig:framework}). We provide hyperparameter sensitivity analysis for $\beta$ and $\tau$ in App. Table \ref{tab:hyperparameter_a}, it shows that our model is not very sensitive to $\beta$ and $\tau$.

\subsection{Learning More Discriminative Features by Boundary Guided Semi-Push-Pull}
\label{sec:bg_sppc}
With the explicit normal and abnormal boundary, we propose a boundary guided semi-push-pull (BG-SPP) loss for more discriminative feature learning. Our BG-SPP loss can utilize the boundary $b_n$ as the contrastive target (boundary guided), and only pull together normal features whose log-likelihoods are smaller than $b_n$ (semi-pull) while pushing abnormal features whose log-likelihoods are larger than $b_a$ apart from $b_n$ at least beyond the margin $\tau$ (semi-push). The formulation of the BG-SPP loss is defined as:
\begin{align}
\label{eq:bg_sppc}
     \mathcal{L}_{bg-spp} = \sum_{i=1}^{N}|&{\rm min}(({\rm log}p_i-b_n),0)| \nonumber \\ &+ \sum_{j=1}^{M}|{\rm max}(({\rm log}p_j - b_n + \tau),0)|
\end{align}
 We define BG-SPP loss as $\ell_1$ norm based formulation to encourage the sparse log-likelihood distribution in the margin region $(b_a,b_n)$, because any log-likelihood ${\rm log}p_i$ fallen into the margin region $(b_a,b_n)$ will increase the value of $\mathcal{L}_{bg-spp}$. Since the log-likelihoods can range from $(-\infty, 0]$, the large region makes it difficult to select the margin hyperparameter $\tau$. Thus, we define a large enough normalizer $\alpha_n$ (\emph{e.g.}, $\alpha_n=10$) and employ it to normalize the log-likelihoods to the range $[-1,0]$. We denote that the extremely small log-likelihoods (less than $-1$) can be excluded outside the BG-SPP loss in E.q.(\ref{eq:bg_sppc}), as these log-likelihoods can be easily divided into anomalies. Therefore, minimizing the BG-SPP loss will encourage all log-likelihoods $\mathcal{P}$ to distribute in the regions $[-1,b_a]$ or $[b_n,0]$. We further analyze the difference between our BG-SPP loss and the hinge loss in Appendix.

In the second training phase, the objective function is as follows:
\begin{equation}
\label{eq:ml_bg_sppc}
    \mathcal{L} = \mathcal{L}_{ml} + \lambda \mathcal{L}_{bg-spp}
\end{equation}


\subsection{Generalization Capability to Unseen Anomalies} 
 \label{sec:generalization_discuss}
 Previous supervised AD methods are usually modeled as binary classification tasks regarding anomalies as positive samples. However, these models rely heavily on the known anomalies. Consequently, these models can overfit the known anomalies, failing to generalize to unseen anomalies. The serious bias issue can be mitigated by our method for three reasons: 1). The obtained explicit separating boundary only relies on the normal feature distribution and has no relation with the abnormal samples, this means that the final decision boundary mainly depends on the normal distribution rather than being affected greatly by the anomalies (see Table \ref{tab:unseen_setting}). 2). Our method still employs the normal distribution to determine anomalies, and our method can form a more compact and discriminative normal feature distribution (this is also conducive to detecting unseen anomalies), rather than the decision boundary between the normals and the known anomalies (see Figure \ref{fig:feature_tsne}). 3). The semi-push-pull mechanism in our method doesn't enforce the anomalies to deviate from the normal distribution as far as possible, which may lead to overfitting of the model for the known anomalies, but only pushes the anomalies outside the margin region (see ablations in Table \ref{tab:semi_push_pull}).
 
\subsection{RandAugment-based Pseudo Anomaly Generation}

Anomalies are generally much less than normal samples, this may cause feature learning inefficient. 
We thus propose a RandAugment-based Pseudo Anomaly Generation (RPAG) strategy, which can simulate anomalies by randomly creating local irregularities, to improve the quantity and diversity of irregular patterns. The whole procedure is shown in Appendix (Figure \ref{fig:data_strategy}) and summarized as follows:

\textbf{Constructing Augmentation Sets.} Adapted from RandAugment \cite{RandAugment}, we first select $K$ available image transformations to construct an augmentation set $\mathcal{T} := \{T_1,\dots,T_K|T_k : \mathcal{I} \rightarrow \mathcal{I}\}$: \{Flip, Rotate, Transpose, Noise, Distortion, Brightness, Sharpness, Translate, Blur\}. 

\textbf{Random Augmentation.} We randomly select an augmentation subset $T_{RS} \subset \mathcal{T}$ containing $S$ transformations to augment an abnormal sample: $(I^a)^{\prime} = T_{RS}(I^a), I^a \in \mathcal{I}^a$. 

\textbf{Selecting Pasting Locations.} Considering that anomalies generally only appear in object regions, thus to guarantee the semantics of simulated anomalies, we should limit the locations of simulated anomalies in the object regions. We adopt a foreground masking strategy, in which we can use grayscale binary thresholding algorithms to effectively locate the foreground objects $M_I = \mathop{Binary}(I^n), I^n \in \mathcal{I}^n$. Then, we can select a random pasting location $r_a = \mathop{Rand}(M_I=1)$ from the object regions ($M_I = 1$) to avoid anomalous regions generated in the background.

\textbf{Cutting Anomalies.} We cut the anomalous regions of the augmented abnormal sample: $\mathcal{R}_a = \mathop{Cut}((I^a)^{\prime})$.

\textbf{Pasting Anomalies.} We paste the cropped anomalous regions back to the normal sample $I^n$ at the selected location $r_a$ to generate a simulated abnormal sample: $I_{sa} = \mathop{Paste}(I^n, \mathcal{R}_a, r_a)$.

  In DRAEM \cite{DRAEM} and NSA \cite{NSA}, the authors also attempt to utilize synthetic anomalies, we ablate our RPAG strategy with their strategies and show results in App. Table \ref{tab:comparison_data_strategy}.
 
  \section{Experiments}
\label{sec:experiments}

\subsection{Datasets and Metrics}

\textbf{Datasets.} In this work, we focus on anomalies in real-world applications, such as industrial defect inspection and medical lesion detection. 
Specifically, we evaluate six real-world anomaly detection datasets, including four industrial defect inspection datasets: \textbf{MVTecAD} \cite{MVTec}, \textbf{BTAD} \cite{BTAD}, \textbf{AITEX} \cite{AITEX} and \textbf{ELPV} \cite{ELPV}; and two medical image datasets for detecting lesions on different organs: \textbf{BrainMRI} \cite{KDAD} and \textbf{HeadCT} \cite{KDAD}. 
A more detailed introduction to these datasets is provided in Appendix.

\textbf{Evaluation Metrics.} For evaluation, the standard metric in anomaly detection, AUROC, is used\cite{MVTec, SSIM, AnoGAN}. Image-level AUROC is used for anomaly detection and a pixel-level AUROC for evaluating anomaly localization. 
In order to weight ground-truth anomaly regions of various sizes equally, we also adopt the Per-Region-Overlap (PRO) curve metric proposed in \cite{STAD}. 

\subsection{Experimental Settings}
\textbf{Multi-Class Setting} is designed to evaluate the performance of AD models in detecting the known anomaly classes. In this setting, the known anomalies are a few abnormal samples randomly drawn from existing anomaly classes in the test set. 
Then, we carefully exclude these added abnormal samples from the test set during testing.

\textbf{One-Class Setting} is designed to evaluate the generalizability of AD models in detecting unseen anomaly classes. In this setting, the known anomalies are randomly sampled only from one anomaly class, and all anomaly samples of this class are removed from the test set to ensure that the test set only contains unseen anomaly classes.

 As anomalies are usually rare, our BGAD is trained with ten random abnormal samples per category by default (we also generate some pseudo anomalies by RPAG based on these known anomalies for training). After removing the known anomalies, the test set in our settings is different from the original test set. Thus, for a fair comparison, we re-run all the compared unsupervised and supervised AD methods with the publicly available implementations under the same experimental setup as our BGAD. Other implementation details can be found in Appendix.

\begin{table*}
\caption{AUROC and PRO results under the multi-class setting on the MVTecAD dataset. $\cdot/\cdot$ means pixel-level AUROC and PRO. The results of our model are averaged over three independent runs. \ddag We implement NFAD as a baseline model, which has the same network structure as our BGAD but without explicit boundary guidance. $*$ We remove the added abnormal samples from the test set, and reproduce all the compared methods under the same experimental setup as our BGAD for a fair comparison.}
\label{tab:main_results}
\resizebox{1.0\linewidth}{!}{
\begin{tabular}{cc|ccccccc|c}
\hline
& \multirow{2}*{Category} &\multicolumn{7}{c|}{Unsupervised AD Methods} &  Supervised AD Method\\
 & & DRAEM$^{*}$ \cite{DRAEM}  & PaDiM$^{*}$ \cite{PaDiM} & MSFD$^{*}$ \cite{MSFD} & PatchCore$^{*}$ \cite{PatchCore} & CFA$^{*}$ \cite{CFA} & NFAD$^{\ddag}$ & BGAD$^{w/o}$ (Ours) & BGAD (Ours) \\
\hline
 \multirow{5}*{\rotatebox{90}{Textures}}& Carpet & 0.954/0.947 & 0.983/0.946 & 0.990/0.958& 0.985/0.959 & 0.989/0.943 & 0.994/0.983  & 0.994/0.982 & \textbf{0.996}$\pm$0.0002/\textbf{0.989}$\pm$0.0004\\
  & Grid  & 0.997/0.984 & 0.963/0.894 & 0.986/0.937 & 0.974/0.891 & 0.977/0.932 & 0.993/0.980  & 0.994/0.980 & \textbf{0.995}$\pm$0.0002/\textbf{0.986}$\pm$0.0001\\
  & Leather & 0.992/0.981 & 0.984/0.966 & 0.978/0.924 & 0.992/0.974 & 0.991/0.958 & 0.997/0.994  & 0.997/0.994 &  \textbf{0.998}$\pm$0.0001/\textbf{0.994}$\pm$0.0003\\
  & Tile & 0.994/0.949 & 0.958/0.884 & 0.952/0.841 & 0.960/0.939 & 0.960/0.860 & 0.969/0.929 & 0.968/0.927 &  \textbf{0.994}$\pm$0.0077/\textbf{0.978}$\pm$0.0021\\
  & Wood  & 0.962/0.935 & 0.963/0.891 & 0.953/0.925  & 0.968/0.857 & 0.948/0.882 & 0.969/0.957 & 0.970/0.957 & \textbf{0.982}$\pm$0.0053/\textbf{0.970}$\pm$0.0007\\
\hline
 \multirow{10}*{\rotatebox{90}{Objects}} & Bottle  & 0.993/0.955 & 0.978/0.936 & 0.985/0.940 & 0.986/0.956 & 0.987/0.944 & 0.988/0.965 & 0.989/0.964 & \textbf{0.994}$\pm$0.0009/\textbf{0.971}$\pm$0.0011\\
 & Cable  & 0.961/0.910 & 0.979/0.973 & 0.972/0.922 & 0.986/\textbf{0.980} & \textbf{0.987}/0.931 & 0.975/0.944  & 0.980/0.968 & 0.986$\pm$0.0010/0.977$\pm$0.0030\\
 & Capsule & 0.869/0.901 & 0.980/0.924 & 0.979/0.878 & 0.990/0.946 & 0.989/0.943 & 0.989/0.952 & 0.992/0.959 & \textbf{0.992}$\pm$0.0021/\textbf{0.964}$\pm$0.0033\\
 & Hazelnut & 0.997/0.985 & 0.980/0.951 & 0.982/0.968  & 0.988/0.924 & 0.986/0.953 & 0.984/0.976 & 0.985/0.976 & \textbf{0.995}$\pm$0.0040/\textbf{0.982}$\pm$0.0028\\
 & Metal nut & 0.992/0.935 & 0.979/0.929 & 0.972/0.985 & 0.986/0.935 & 0.987/0.918 & 0.971/0.942  & 0.976/0.948 &  \textbf{0.996}$\pm$0.0003/\textbf{0.970}$\pm$0.0012\\
 & Pill & 0.979/0.959 & 0.978/0.957 & 0.971/0.929 & 0.983/0.947 & 0.986/0.965 & 0.976/0.978  & 0.980/0.980 & \textbf{0.996}$\pm$0.0002/\textbf{0.988}$\pm$0.0005\\
 & Screw & 0.992/0.965 & 0.974/0.923 & 0.983/0.924 & 0.984/0.928 & 0.985/0.944 & 0.988/0.945 & 0.992/0.960 & \textbf{0.993}$\pm$0.0003/\textbf{0.968}$\pm$0.0010\\
 & Toothbrush & 0.970/0.940 & 0.980/0.894 & 0.986/0.877 & 0.987/0.939 & 0.989/0.894 & 0.983/0.904 & 0.986/0.938 & \textbf{0.995}$\pm$0.0003/\textbf{0.961}$\pm$0.0026\\
 & Transistor & 0.970/0.935 & 0.983/0.967 & 0.886/0.781 & 0.964/0.967 & \textbf{0.985}/0.960 & 0.923/0.788  & 0.940/0.830 & 0.983$\pm$0.0005/\textbf{0.972}$\pm$0.0015\\
 & Zipper & 0.984/0.966 & 0.978/0.948 & 0.981/0.935 & 0.986/0.963 & 0.988/0.944 & 0.986/0.957 & 0.987/0.957 & \textbf{0.993}$\pm$0.0003/\textbf{0.977}$\pm$0.0002\\
\hline
\hline
 & \textbf{Mean} & 0.969/0.947 & 0.976/0.932 & 0.970/0.915 & 0.981/0.940 & 0.982/0.931 & 0.979/0.946 & 0.982/0.955 &  \textbf{0.992}$\pm$0.0007/\textbf{0.976}$\pm$0.0006\\
\hline
 & \textbf{Image-level Mean} & 0.978 & 0.975 & 0.964 & 0.988 & 0.989 & 0.968  & 0.974 & \textbf{0.993}$\pm$0.0012\\
\hline
\end{tabular}}
\end{table*}

\begin{table}
\caption{AUROC and PRO results under the multi-class setting on the MVTecAD dataset. $*$ Please see explanation in Table \ref{tab:main_results}.}
\label{tab:main_sup_results}
\resizebox{1.0\linewidth}{!}{
\begin{tabular}{c|cccc}
\hline
\multirow{2}*{Category} & \multicolumn{4}{c}{Supervised AD Methods (Ten Abnormal Samples)} \\
 & FCDD$^{*}$ \cite{FCDD} & DevNet$^{*}$ \cite{DevNet} & DRA$^{*}$ \cite{DRA} & BGAD (Ours) \\
\hline
  Carpet & 0.981/0.952 & -/- & -/- & \textbf{0.996}$\pm$0.0002/\textbf{0.989}$\pm$0.0004\\
  Grid  & 0.949/0.897 & -/- & -/- & \textbf{0.995}$\pm$0.0002/\textbf{0.986}$\pm$0.0001\\
  Leather & 0.984/0.973 & -/- & -/- &  \textbf{0.998}$\pm$0.0001/\textbf{0.994}$\pm$0.0003\\
  Tile & 0.977/0.938 & -/- & -/- &  \textbf{0.994}$\pm$0.0077/\textbf{0.978}$\pm$0.0021\\
  Wood  & 0.950/0.901 & -/- & -/- & \textbf{0.982}$\pm$0.0053/\textbf{0.970}$\pm$0.0007\\
\hline
 Bottle  & 0.966/0.939 & -/- & -/- & \textbf{0.994}$\pm$0.0009/\textbf{0.971}$\pm$0.0011\\
 Cable  & 0.963/0.980 & -/- & -/- & 0.986$\pm$0.0010/0.977$\pm$0.0030\\
 Capsule & 0.970/0.922 & -/- & -/- & \textbf{0.992}$\pm$0.0021/\textbf{0.964}$\pm$0.0033\\
 Hazelnut & 0.970/0.958 & -/- & -/- & \textbf{0.995}$\pm$0.0040/\textbf{0.982}$\pm$0.0028\\
 Metal nut & 0.966/0.934 & -/- & -/- &  \textbf{0.996}$\pm$0.0003/\textbf{0.970}$\pm$0.0012\\
  Pill & 0.975/0.960 & -/- & -/- & \textbf{0.996}$\pm$0.0002/\textbf{0.988}$\pm$0.0005\\
 Screw & 0.963/0.925 & -/- & -/- & \textbf{0.993}$\pm$0.0003/\textbf{0.968}$\pm$0.0010\\
 Toothbrush & 0.967/0.907 & -/- & -/- & \textbf{0.995}$\pm$0.0003/\textbf{0.961}$\pm$0.0026\\
 Transistor & 0.942/0.935 & -/- & -/- & 0.983$\pm$0.0005/\textbf{0.972}$\pm$0.0015\\
 Zipper & 0.968/0.948 & -/- & -/- & \textbf{0.993}$\pm$0.0003/\textbf{0.977}$\pm$0.0002\\
\hline
\hline
 \textbf{Mean} & 0.966/0.938 & -/- & -/- &  \textbf{0.992}$\pm$0.0007/\textbf{0.976}$\pm$0.0006\\
\hline
 \textbf{Image-level Mean} & 0.965 & 0.948 & 0.961 & \textbf{0.993}$\pm$0.0012\\
\hline
\end{tabular}}
\end{table}

\subsection{Results under the Multi-Class Setting}

\textbf{MVTecAD.} We compare our BGAD with the SOTA AD methods, including unsupervised (PaDiM \cite{PaDiM}, DRAEM \cite{DRAEM}, MSFD \cite{MSFD}, PatchCore \cite{PatchCore} and CFA \cite{CFA}) and supervised methods (FCDD \cite{FCDD}, DevNet\cite{DevNet} and DRA \cite{DRA}). The detailed comparison results of all categories are shown in Table \ref{tab:main_results} and Table \ref{tab:main_sup_results}. We also implement a variant BGAD$^{w/o}$, which is optimized by the first part of the BG-SPP loss without anomalies. Compared with NFAD, our BGAD$^{w/o}$ can achieve better results, this shows that our boundary guiding mechanism is also beneficial to improve the AD performance under the unsupervised setting. Our BGAD reaches the best performance under all three evaluation metrics and can further surpass unsupervised baseline NFAD by 2.5\% and 1.3\% AUROC, and 3.0\% PRO. The largest gain in PRO demonstrates that our BGAD is more suitable for anomaly localization to better locate the anomalous areas (see Figure \ref{fig:qualitative_results}). Compared with supervised AD methods, our method can also surpass these SOTA methods significantly. This shows that our boundary guiding mechanism can exploit a few known anomalies more effectively to learn a more discriminative AD model. 

\textbf{BTAD.} We compare our BGAD with three baseline methods reported in \cite{BTAD}: AE-MSE, AE-SSIM, and VT-ADL. Following \cite{BTAD}, we evaluate anomaly localization performance and report pixel-level AUROCs. The results are shown in Table \ref{tab:BTAD}. Our BGAD can achieve 98.6\% mean pixel-level AUROC, which surpasses other methods by a large margin (8.6\%) and surpasses unsupervised baseline NFAD by 0.8\%. What's more, BGAD can achieve 82.4\% PRO which surpasses unsupervised NFAD by 4.6\%.

\begin{table}[t]
    \caption{Pixel-level AUROC results on the BTAD dataset. $\cdot/\cdot$ means pixel-level AUROC and PRO. The results of our model are averaged over three independent runs.}
\label{tab:BTAD}
\resizebox{\linewidth}{!} {
\begin{tabular}{c|c|c|c|c|c}
\hline
Categories & AE-MSE  & AE-SSIM  & VT-ADL & NFAD & BGAD (Ours) \\
\hline
1 & 0.490 & 0.530 & \textbf{0.990} & 0.972/0.767 & 0.982$\pm$0.0027/\textbf{0.830}$\pm$0.0318 \\
\hline
2 & 0.920 & 0.960 & 0.940 & 0.967/0.578 & \textbf{0.979}$\pm$0.0018/\textbf{0.648}$\pm$0.0173\\
\hline
3 & 0.950 & 0.890 & 0.770 & 0.996/0.988 & \textbf{0.998}$\pm$0.0003/\textbf{0.993}$\pm$0.0005\\
\hline
Mean & 0.780 & 0.790 & 0.900 & 0.978/0.778 & \textbf{0.986}$\pm$0.0015/\textbf{0.824}$\pm$0.0163\\
\hline
\end{tabular}}
\end{table}

\begin{figure}[ht]
    \centering
    \includegraphics[width=1.0\linewidth]{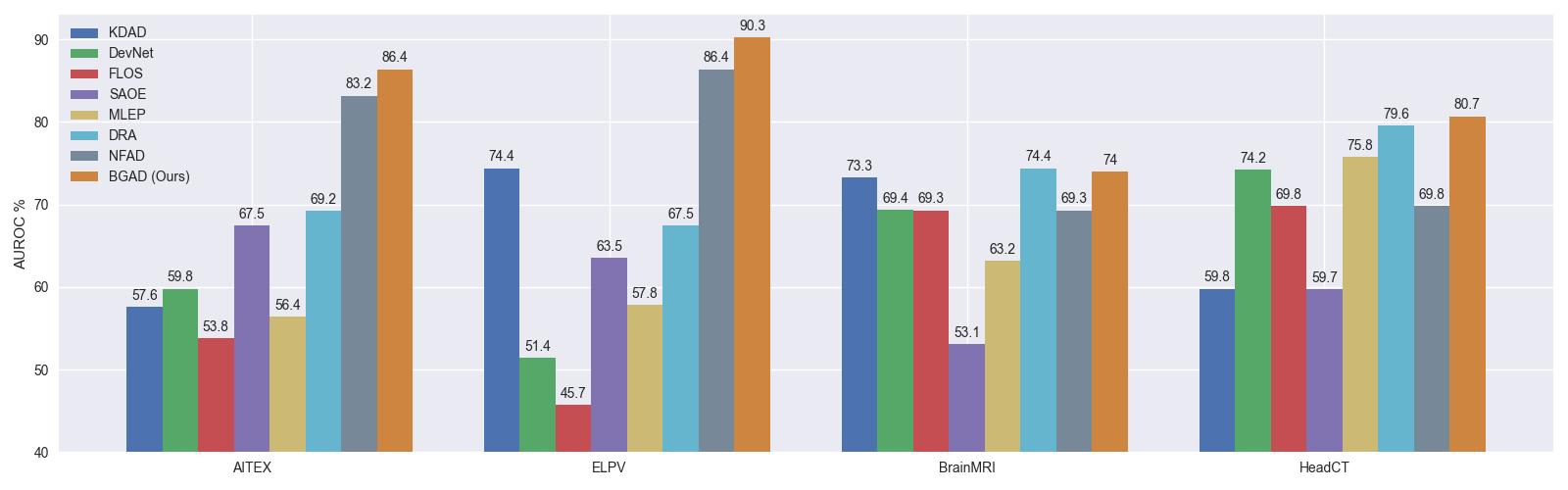}
    \caption{AUROC results on the AITEX, ELPV, BrainMRI, and HeadCT datasets.}
    \label{fig:other_dataset}
\end{figure}

\textbf{Other Datasets.} Here, we compare our BGAD with six recent and closely related SOTA methods reported in \cite{DRA}: unsupervised KDAD \cite{MSFD}, and supervised DevNet \cite{DevNet}, FLOS \cite{FocalLoss}, SAOE \cite{SAOE}, MLEP \cite{MLEP} and DRA \cite{DRA}. Following \cite{DRA}, we evaluate anomaly detection performance and report image-level AUROCs. Same as \cite{DRA}, all models are trained with one known anomaly sample. The comparison results are shown in Figure \ref{fig:other_dataset}. Our model can achieve the best AUROC performance on the two industrial defect inspection datasets (AITEX and ELPV), and comparable results with the SOTA methods on the two medical lesion detection datasets (BrainMRI and HeadCT).



\subsection{Results under the One-Class Setting}
The comparison results under the one-class setting are shown in Table \ref{tab:unseen_setting}, and more results are shown in Table \ref{tab:unseen_setting_append} in Appendix.

\textbf{Comparison to Supervised AD Methods.} In both Table \ref{tab:unseen_setting} and Table \ref{tab:unseen_setting_append}, compared to the competing methods, our method is the best performer on the diverse application datasets. On the mean image-level performance, our model achieves about 3.3\%-12.9\% mean AUROC increase compared to the best contender. This shows substantially better generalizability of our model in detecting unseen anomaly classes than the other supervised AD methods. 

\textbf{Comparison to Unsupervised Baseline.} Since supervised AD methods are often biased by the seen anomaly class, they even perform less effectively than the unsupervised baseline NFAD on most of the datasets. By contrast, our model can outperform the baseline NFAD across all the datasets. The comparison results validate that our model's better generalizability to unseen anomalies and the serious bias issue can be alleviated by our method.

\begin{table}
\caption{AUROC results under the one-class setting, where models are trained with only one anomaly class and tested to detect other anomaly classes. $\cdot/\cdot$ means image-level and pixel-level AUROCs.}
\label{tab:unseen_setting}
\resizebox{1.0\linewidth}{!}{
\begin{tabular}{cc|c|cccccc}
\hline
 & \multirow{2}*{Known Class} & Baseline & \multicolumn{6}{c}{Ten Training Anomaly Samples} \\
 \cline{3-9}
 & & NFAD & DevNet  & FLOS  & SAOE  & MLEP  & DRA  & BGAD (Ours)\\
\hline
 \multirow{6}*{\rotatebox{90}{Carpet}} & Color & 0.998/\textbf{0.993} & 0.767/- & 0.760/-  & 0.467/- & 0.689/- & 0.886/- & \textbf{1.000}/\textbf{0.993} \\
  & Cut & \textbf{0.998}/0.995 & 0.819/- & 0.688/-  & 0.793/- & 0.653/- & 0.922/- & \textbf{0.998}/\textbf{0.996} \\
  & Hole & 0.997/0.993 & 0.814/- & 0.733/-  & 0.831/- & 0.674/- & 0.922/- &  \textbf{0.998}/\textbf{0.995} \\
  & Metal & 0.998/0.993  & 0.863/- & 0.678/- & 0.883/- & 0.764/- & 0.933/- &  \textbf{1.000}/\textbf{0.994} \\
  & Thread & \textbf{1.000}/0.995 & 0.972/- & 0.946/- & 0.834/- & 0.967/- & 0.989/- & \textbf{1.000}/\textbf{0.996} \\
  \cline{2-9}
  & \textbf{Mean} & 0.998/0.994 & 0.847/- & 0.761/- & 0.762/- & 0.751/- & 0.935/- & \textbf{0.999}/\textbf{0.995} \\
\hline
 \multirow{5}*{\rotatebox{90}{Metal\_nut}} & Bent & 0.977/0.959 & 0.904/- & 0.827/- & 0.901/- & 0.956/- & 0.990/- & \textbf{1.000}/\textbf{0.972}\\
 & Color & 0.977/0.963 & 0.978/- & 0.9788/-  & 0.879/- & 0.945/- & 0.967/- & \textbf{0.999}/\textbf{0.973} \\
 & Flip & 0.976/0.977 & 0.987/- & 0.942/- & 0.795/-  & 0.805/- & 0.913/- & \textbf{0.995}/\textbf{0.982}\\
 & Scratch & \textbf{1.000}/0.965 & 0.991/- & 0.943/- & 0.845/- & 0.805/- & 0.911/- & \textbf{1.000}/\textbf{0.972} \\
 \cline{2-9}
 & \textbf{Mean} & 0.983/0.966 & 0.965/- & 0.922/-  & 0.855/- & 0.878/- & 0.945/- &  \textbf{0.998}/\textbf{0.975}\\
\hline
\hline
\end{tabular}}
\end{table}

\subsection{Ablation Study}
\label{sec:ablation}
 
 \textbf{Experiments On Hard Subsets.} The experimental results in Table \ref{tab:main_results} has already demonstrated the effectiveness of our model. However, to further demonstrate the ability of our method to detect complex anomalies, we construct two more difficult subsets from the MVTecAD dataset and conduct experiments on these two subsets. The details of subset selection are provided in Appendix. 
 The results are shown in Table \ref{tab:harder_subset}. It can be found that the detection and localization performance gain on these hard subsets is larger than that on the original dataset with a margin of 1.1\%, 1.3\%, and 5.6\% respectively. This ablation study demonstrates that our model is more beneficial for harder anomaly classes.
 
 \begin{table}[t]
\centering
\caption{AUROC and PRO results on subsets from the MVTecAD dataset. The details of subset selection are provided in Appendix.}
\label{tab:harder_subset}
\resizebox{\linewidth}{!} {
\begin{tabular}{c|c|c|c|c|c|c}
\hline
\multirow{2}*{\diagbox{Metric}{Dataset}} & \multicolumn{2}{c|}{MVTecAD} & \multicolumn{2}{c|}{Hard Subsets} & \multicolumn{2}{c}{Unseen Subsets} \\
\cline{2-7}
& NFAD & BGAD & NFAD & BGAD & NFAD & BGAD\\
\hline
Image AUROC & 0.968 & 0.992(+2.5\%) & 0.948 & 0.984(\textbf{+3.6\%}) & 0.948 & 0.971(+\textbf{2.3}\%)\\
\hline
Pixel AUROC & 0.979 & 0.992(+1.3\%) & 0.960 & 0.986(\textbf{+2.6\%}) & 0.960 & 0.982(+\textbf{2.2}\%)\\
\hline
PRO & 0.946 & 0.976(+3.0\%) & 0.863 & 0.949(\textbf{+8.6\%}) & 0.863 & 0.930(+\textbf{6.7}\%)\\
\hline
\end{tabular}}
\end{table}

\textbf{Generalization to Hard Subsets.} We use the easy subsets as the training set and validate results on the hard subsets to explore the generalizability of the model. The easy subsets are formed by excluding the hard subsets mentioned in the last paragraph from the original dataset. The experimental results are shown in Table \ref{tab:harder_subset}. It can be found that even only trained with easy anomalies, our BGAD can generalize well to hard anomalies with performance gain by 2.3\% and 2.2\% AUROC, and 6.7\% PRO.

\textbf{Effect of Semi-Push-Pull Mechanism.} We implement a variant of BGAD (termed as BGAD$^{\dag}$), it doesn't utilize the BG-SPP loss while employing the conventional contrastive loss. The comparison results are shown in Table \ref{tab:semi_push_pull}, and more results are in Appendix (Table \ref{tab:semi_push_pull_append}). The BGAD$^{\dag}$ performs less effectively than the BGAD, and even worse than the baseline NFAD (especially for some complex categories, \emph{e.g.}, Capsule, Screw, Transistor). The reason is that the BGAD$^{\dag}$'s full pushing mechanism will encourage the anomalous features to deviate from the normal distribution at least a large enough bound, which may make the model more inclined to generate larger anomaly scores and thus lead to the model being easier to over-fit the known anomalies. Therefore, the BGAD$^{\dag}$ may generate larger anomaly scores for normal features, which will significantly reduce the AUROC metrics. However, the semi-push-pull mechanism in our BGAD only changes the ambiguous region, this has less impact on the full normal and abnormal distributions. Thus, the BG-SPP loss doesn't make the model have the inclination to generate larger anomaly scores, is more conducive to alleviating over-fitting of the model to the known anomalies.


\begin{table}
\caption{AUROC results under the one-class setting. $\cdot/\cdot$ means image-level and pixel-level AUROCs.}
\label{tab:semi_push_pull}
\resizebox{1.0\linewidth}{!}{
\begin{tabular}{c|ccccc}
\hline
 \multirow{2}*{Method} & \multicolumn{5}{c}{Category} \\
 \cline{2-6}
 & Carpet & Metal\_nut & Capsule  & Screw & Transistor \\
\hline
 NFAD (baseline) & 0.998/0.994 & 0.983/0.966 & 0.941/0.990 & 0.885/0.989 & 0.984/0.929 \\
  BGAD$^{\dag}$ & 0.998/0.994 & 0.997/0.927 & 0.868/0.963 & 0.823/0.980 & 0.933/0.847 \\
  BGAD (Ours) & \textbf{0.999}/\textbf{0.995} & \textbf{0.998}/\textbf{0.975} & \textbf{0.988}/\textbf{0.991} & \textbf{0.947}/\textbf{0.991} & \textbf{0.994}/\textbf{0.942} \\
\hline
\hline
\end{tabular}}
\end{table}
 
 \subsection{Qualitative Results}
\begin{figure}[ht]
    \centering
    \includegraphics[width=1.0\linewidth]{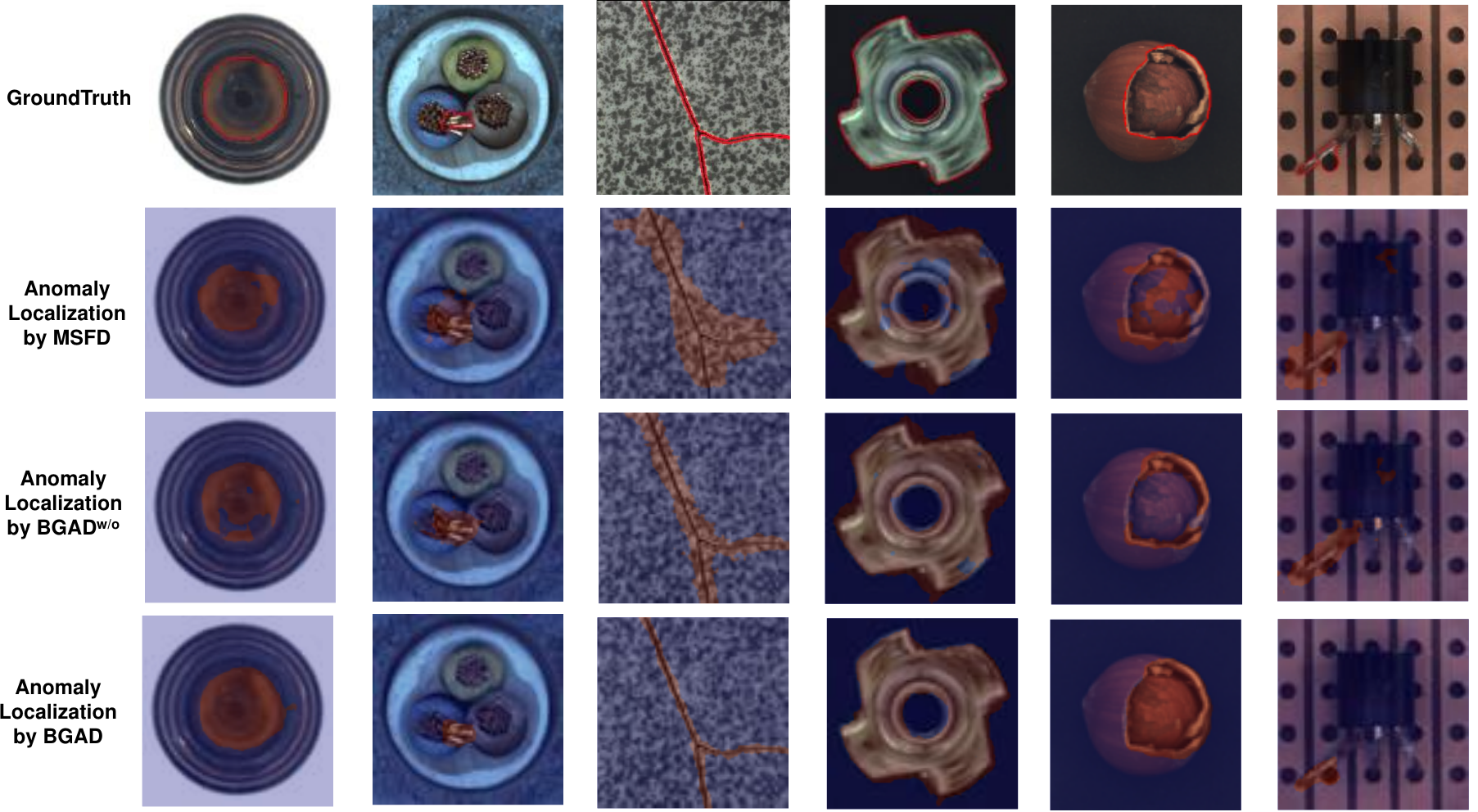}
    \caption{Qualitative results. The anomaly localization results generated by MSFD, BGAD$^{w/o}$, and BGAD are shown for comparison. In the first row, the areas enclosed by the red lines are ground-truth.}
    \label{fig:qualitative_results}
\end{figure}

We visualize some anomaly localization results in Figure \ref{fig:qualitative_results} with the MVTecAD dataset. Our BGAD can generate more accurate anomaly localization maps (see columns of \{1,3,4,5,6\} in Figure \ref{fig:qualitative_results}), or even generate anomaly maps better than ground truth (see columns of \{2\} in Figure \ref{fig:qualitative_results}). 

\begin{figure}[ht]
    \centering
    \includegraphics[width=1.0\linewidth]{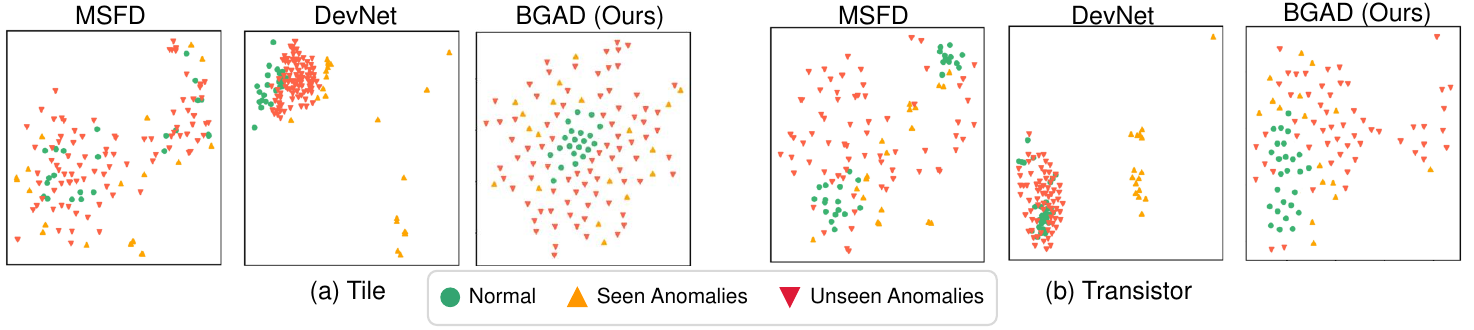} 
    \caption{Feature distributions learned by unsupervised MSFD \cite{MSFD}, supervised DevNet \cite{DevNet} and our BGAD.}
    \label{fig:feature_tsne}
\end{figure}

\begin{figure}[ht]
    \centering
    \includegraphics[width=1.0\linewidth]{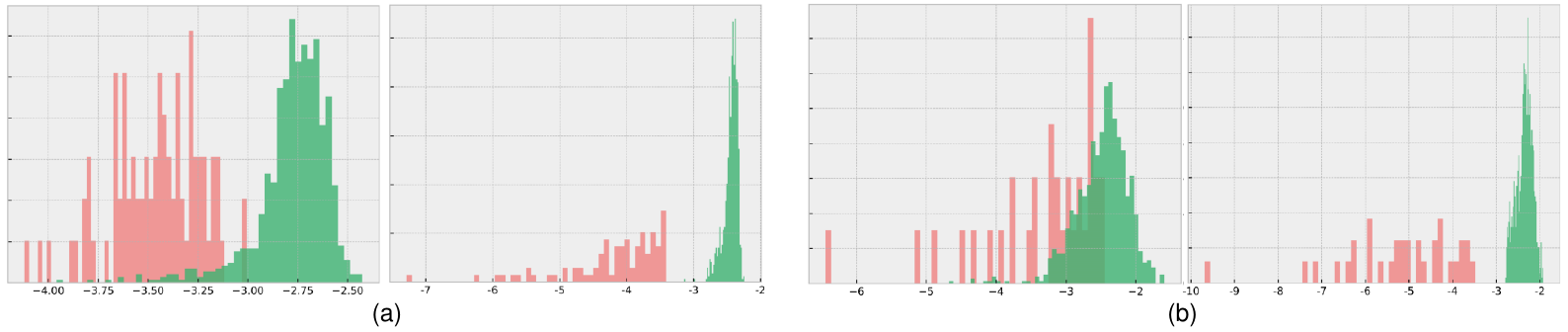}
    \caption{Log-likelihood histograms from (a) tile and (b) transistor category. Left is the log-likelihood histogram w/o anomaly samples, right is the log-likelihood histogram with anomaly samples.}
    \label{fig:histgram}
\end{figure}

To illustrate the effectiveness of our method more intuitively, we visualize normal and abnormal feature distributions and log-likelihood distributions in Figure \ref{fig:feature_tsne}, \ref{fig:histgram}. From Figure \ref{fig:feature_tsne}, it can be found that the supervised DevNet \cite{DevNet} is biased by the known anomalies, failing to distinguish unseen anomalies from the normal data. But our method can effectively mitigate this issue and generate more discriminative features than the unsupervised MSFD \cite{MSFD}. From Figure \ref{fig:histgram}, it can be found that the ambiguous log-likelihood regions can be diminished by our BGAD.


\section{Conclusion}
\label{sec:conclusion}
We propose a novel and more discriminative AD model termed as BGAD to tackle the \emph{insufficient discriminability} issue and the \emph{bias} issue simultaneously. Compared with unsupervised AD models, our model can learn more discriminative features for distinguishing anomalies by exploiting a few anomalies effectively. Compared with supervised AD methods, our method can mitigate the bias issue with the explicit separating boundary and semi-push-pull mechanism. We hope our boundary guiding mechanism can inspire subsequent studies of supervised AD.

\section*{Acknowledgements}
This work was supported in part by the National Natural Science Fund of China (61971281), the Shanghai Municipal Science and Technology Major Project (2021SHZDZX0102), and the Science and Technology Commission of Shanghai Municipality (22DZ2229005).

{\small
\bibliographystyle{ieee_fullname}
\bibliography{egbib}
}

\clearpage
\section*{Appendix}
\appendix
\label{sec:appendix}

\section{Limitations and Future Work}
\label{sec:appendix_limitations}
In this paper, we propose a novel and more discriminative anomaly detection model termed as BGAD to tackle the \emph{insufficient discriminability} issue and the \emph{bias} issue simultaneously. But there are still some limitations to our method. Here, we discuss two main limitations as follows:

One limitation of our model is that we employ normalizing flow to obtain the explicit separating boundary due to its exact log-likelihood estimation ability. However, not all anomaly detection models can generate log-likelihoods, thus our boundary guiding mechanism can't be used in these models directly. A possible solution is that we can use pairwise distances (vector dot product of two features) to substitute log-likelihoods, and then obtain the explicit boundary based on the distribution of normal pairwise distances, and then use BG-SPP loss to optimize the model to learn more discriminative features. 

Another limitation is that our method requires anomalous samples to achieve better results, but it is difficult to collect all kinds of anomalies. Thus, generalization performance to unseen anomalies is a critical problem that we should consider, the experimental results in Table \ref{tab:unseen_setting}, \ref{tab:harder_subset} and \ref{tab:unseen_setting_append} validate our model's generalization capability. However, further improving our model's generalizability and theoretical analysis of model's generalizability are still important and valuable future works. Future work also includes tackling the imbalance problem between normal and abnormal more effectively and attempting to only use pseudo anomalies in our method, such as the generated pseudo anomalies in these works \cite{CutPaste, DRAEM, DRA}.

\section{Dataset Details}
\label{sec:appendix_dataset}
\textbf{MVTecAD.} The MVTec Anomaly Detection dataset \cite{MVTec} contains 5354 high-resolution images (3629 images for training and 1725 images for testing) of 15 different categories. 5 classes consist of
textures and the other 10 classes contain objects. A total of 73 different defect types are presented and almost 1900 defective regions are manually annotated in this dataset.

\textbf{BTAD.} The BeanTech Anomaly Detection dataset \cite{BTAD} contains 2830 real-world images of 3 industrial products. Product 1, 2, and 3 of this dataset contain 400, 1000, and 399 training images respectively.

\textbf{AITEX.} The AITEX \cite{AITEX} is a fabric defect inspection dataset that has 12 defect categories. The original images in this dataset are $4096\times 256$ resolution, we convert this dataset to MVTecAD format following the converting method used in the work \cite{DRA}.

\textbf{ELPV.} The ELPV \cite{ELPV} dataset contains 2642 samples of $300\times 300$ resolution. The dataset is used for solar cell defect inspection and contains two defect categories: mono- and poly-crystalline.

\textbf{BrainMRI}. The BrainMRI is a brain tumor detection dataset obtained by magnetic resonance imaging (MRI) of the brain. The dataset can be downloaded from the Kaggle competition \url{https://www.kaggle.com/datasets/navoneel/brain-mri-images-for-brain-tumor-detection}.

\textbf{HeadCT}. The HeadCT is a head hemorrhage detection dataset obtained by a CT scan of the head. The dataset can be downloaded from the Kaggle competition \url{https://www.kaggle.com/datasets/felipekitamura/head-ct-hemorrhage}.

\section{Implementation Details}
\label{sec:appendix_imp}
As illustrated in Figure \ref{fig:framework}, we use Efficient-b6 \cite{Efficientnet} pre-trained on ImageNet \cite{timm} dataset as the feature extractor to extract three levels of feature maps with $\{4\times, 8\times, 16\times\}$ downsampling ratios. The parameters of the feature extractor are frozen in the training process, only the parameters of the normalizing flow are learnable. The extracted multi-scale features are then transformed to latent space by the normalizing flow, the normalizing flow is constructed by $8$ coupling layers similar to \cite{CFLOW}. We train BGAD and BGAD$^{w/o}$ using Adam optimizer with $2e-4$ learning rate, 200 train epochs, 32 mini-batch size, and cosine learning rate annealing strategy with $2$ warm-up epochs. The normalizer described in sec \ref{sec:bg_sppc} is set to 10 by default, the hyperparameter $\lambda$ in E.q.(\ref{eq:ml_bg_sppc}) is set to 1.0 by default, the default number of transformations in augmentation subset is set as 3. The hyperparameter $\beta$ is set to 1 by default, and the $\tau$ is set to 0.1 by default. With the default hyperparameters, our BGAD can achieve effective performance improvement over NFAD on the six datasets. All the training and test images are resized and cropped to $256\times256$ resolution from the original resolution. We also utilize a balanced batch sampler to ensure that the ratio of normal and abnormal samples in each mini-batch is 2:1, which can mitigate the rarity problem at the batch level. Our main code is based on the CFLOW implementation made public by the authors of \cite{CFLOW} under the MIT license. The pre-trained feature extractor Efficient-b6 is from the timm \cite{timm} library under the Apache 2.0 license.

\begin{figure}
    \includegraphics[width=1.0\linewidth]{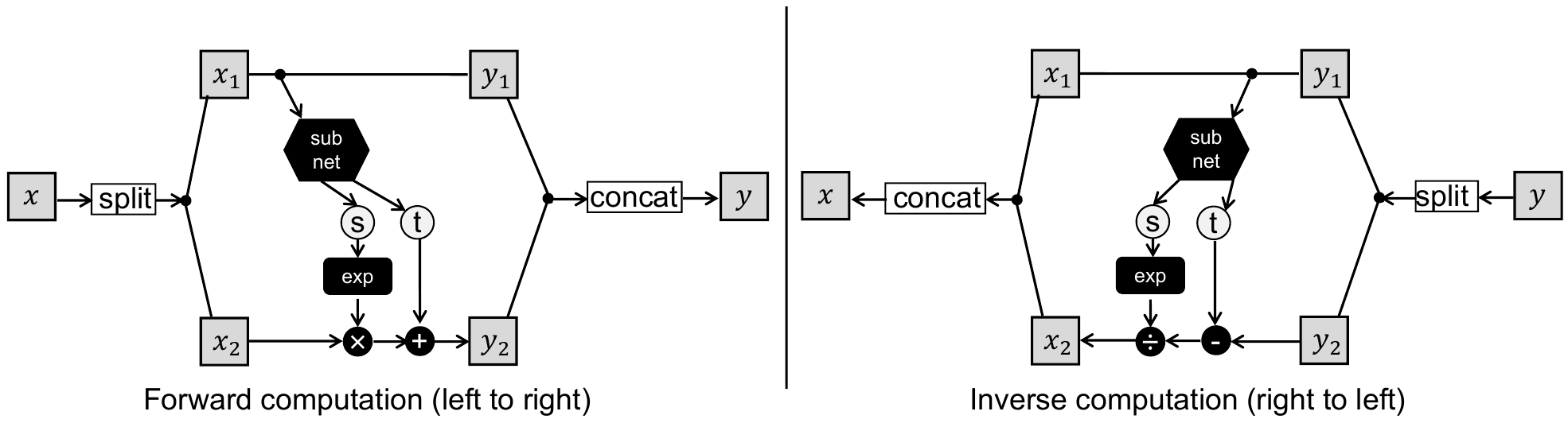} \\
    \caption{Illustration of a coupling layer. The transformation coefficients are predicted by a subnetwork (\emph{subnet}), which is composed of fully-connected layers, nonlinear activations, batch normalization layers, \emph{etc.}. The forward affine coupling can be calculated as $y_1 = x_1; y_2 = x_2 \odot {\rm exp}(s(x_1)) + t(x_1)$, and the inverse affine coupling is then calculated as $x_1 = y_1; x_2 = (y_2 - t(y_1)) / {\rm exp}(s(y_1))$.}
    \label{fig:affine_coupling}
\end{figure}

The normalizing flow in our model is mainly based on Real-NVP \cite{realNVP} architecture, but the convolutional subnetwork in Real-NVP is replaced with a linear subnetwork. Our CNFlow also combines many design efforts of various works on normalizing flows from recent works \cite{SoftPerm, Glow, i-revnet}. As in previous works, the normalizing flow in our model is composed of the so-called \emph{coupling layers}. In our CNFlow, each coupling layer is designed to achieve the forward or inverse affine coupling computation \cite{realNVP} as illustrated in Figure \ref{fig:affine_coupling}. And the native coupling layer is followed by random and fixed soft permutation of channels \cite{SoftPerm}, and a fixed scaling by a constant, similar to ActNorm layers introduced by \cite{Glow}. For the coupling coefficients, each subnetwork predicts multiplicative and additive components jointly, as done by \cite{Glow}. Furthermore, we adopt the soft clamping of multiplication coefficients used by \cite{realNVP}. The implementation of the normalizing flow in our model is based on the FrEIA library \url{https://github.com/VLL-HD/FrEIA}, thanks to the authors' contributions, we can implement non-trivial normalizing flow conveniently.

The total parameters of our model are 43M, but the learnable parameters are only 3.6M. Our model can be trained by one GPU card, and the memory usage of our model is only about 2600MB, which means that training our model generally doesn't appear out-of-memory issue. In experiments, we train the model on the MVTecAD dataset with one Titan-XP GPU card, the total training time is about 30 hours, and the inference speed of our model is about 16fps achieved by Titan-XP. Code will be publicly available online.

\section{Error Bound Analysis}
\label{sec:appendix_th1}
\textbf{Proposition 1.} \emph{Assume that $\varphi_{\theta^*} \in {\rm argmin}_{\varphi_\theta, \theta \in \Theta}\{\mathcal{L}_{ml}+\lambda\mathcal{L}_{bg-spp}\}$, and $y=0$, $y=1$ means normal and abnormal features. Then we have that
\begin{align}
    & \mathbb{E}_{y_i=0}[{\rm max}((b_n^\prime-{\rm log}p_i),0)]+\mathbb{E}_{y_j=1}[{\rm max}(({\rm log}p_j-b_a^\prime),0)] \nonumber \\
    & \leq (b_n-b_a)\mathcal{L}_{bg-spp}(\varphi_{\theta^*}) + N/(N+M)[{\rm max}(1+b_n^\prime,-b_a^\prime)] \nonumber \\
    & \leq \frac{(\frac{d}{2}{\rm log}(2\pi)-\frac{1}{2})(b_n - b_a)}{\lambda} + N/(N+M)
\end{align}
where the $b_n^\prime=b_n - \epsilon, b_a^\prime = b_a + \epsilon, \epsilon \in (0,b_n - b_a)$, $N$ and $M$ are the number of normal and abnormal features.}

\emph{proof.} The derivation procedure mainly follows the Theorem 3 in \cite{LargeMargin}. We denote that ${\rm log}p_1 \geq {\rm log}p_2 \geq \dots \geq {\rm log}p_{N+M}$ is a ranking of the log likelihoods, where $N$ and $M$ are the number of normal and abnormal log likelihoods respectively. Then for $b_n = {\rm log}p_{N}$, we have that
\begin{align}
\label{eq:theorem1}
     & \mathbb{E}_{y_i=0}[{\rm max}((b_n^\prime-{\rm log}p_i),0)]+\mathbb{E}_{y_j=1}[{\rm max}(({\rm log}p_j-b_a^\prime),0)] \nonumber \\
     & \leq (b_n^\prime-b_a^\prime)\mathcal{L}_{bg-spp}^{0}(\varphi_{\theta^*}) + N/(N+M)[{\rm max}(1+b_n^\prime,-b_a^\prime)] \nonumber \\
     & \leq (b_n-b_a)\mathcal{L}_{bg-spp}(\varphi_{\theta^*}) + N/(N+M) 
\end{align}
where the $\mathcal{L}_{bg-spp}^{0}$ means the $\ell_0$ norm based formulation of the BG-SPP loss. The first inequality is obtained by assuming the worst case where ${\rm log}p_1,\dots,{\rm log}p_{N}$ are all misclassified and the others are fallen in $(b_a,b_n)$. The second inequality is obtained as $1+b_n^\prime \leq 1$ and $-b_a^\prime \leq 1$ when $-1 \leq b_a^\prime < b_n^\prime \leq 0$ satisfies.

Furthermore, for any $\varphi_0$ satisfying $\mathcal{L}_{bg-spp}(\varphi_0)=0$, by the optimality of $\varphi_{\theta*}$, we have that
\begin{align}
    \mathcal{L}_{ml}(\varphi_{\theta^*}) +\lambda\mathcal{L}_{bg-spp}(\varphi_{\theta^*})&\leq \mathcal{L}_{ml}(\varphi_{\theta^0})+\lambda\mathcal{L}_{bg-spp}(\varphi_{\theta^0}) \nonumber \\ &=\mathcal{L}_{ml}(\varphi_{\theta^0})
\end{align}
and thus (the similarity function $g$ in BG-SPP loss is specified as the general exponentiated-cosine distance function for simplifying derivation) 
\begin{align}
\label{eq:theorem1_prove}
    & \mathcal{L}_{bg-spp}(\varphi_{\theta^*}) \nonumber \\
    & \leq (\mathcal{L}_{ml}(\varphi_{\theta^0}) -  \mathcal{L}_{ml}(\varphi_{\theta^*})) / \lambda  \nonumber \\
    & \leq \frac{1}{\lambda} \Bigg(\frac{1}{2}\varphi_{\theta^0}(x)^T\varphi_{\theta^0}(x)-\frac{1}{2}\varphi_{\theta^*}(x)^T\varphi_{\theta^*}(x) \nonumber \\ &+\frac{1}{2}\varphi_{\theta^*}(x)^T\varphi_{\theta^*}(x)+\frac{d}{2}{\rm log}(2\pi)\Bigg) \nonumber \\
    & \leq \frac{1}{\lambda}\Bigg(-\frac{1}{2} + \frac{d}{2}{\rm log}(2\pi)\Bigg) \nonumber \\
    & = \frac{\frac{d}{2}{\rm log}(2\pi) - \frac{1}{2}}{\lambda}
\end{align}
The second inequality is obtained by assuming the worst initial states:
\begin{equation}
     \varphi_{\theta^0}(x)^T\varphi_{\theta^0}(x) = -1 
\end{equation}
By combining the above E.q.(\ref{eq:theorem1}) and E.q.(\ref{eq:theorem1_prove}), we have that
\begin{align}
    & \mathbb{E}_{y_i=0}[{\rm max}((b_n^\prime-{\rm log}p_i),0)]+\mathbb{E}_{y_j=1}[{\rm max}(({\rm log}p_j-b_a^\prime),0)] \nonumber \\
    & \leq \frac{(\frac{d}{2}{\rm log}(2\pi)-\frac{1}{2})(b_n - b_a)}{\lambda} + N/(N+M)
\end{align}

The above proposition demonstrates that the necessity and usefulness of the BG-SPP loss, because increasing the hyperparameter $\lambda$ would assist the error bound in converging to zero. And the proposition also implies that increasing anomalies will benefit the reliability of the normal and abnormal discrimination. We also empirically validate the effect of $\lambda$ on the detection results in Table \ref{tab:labmda}, the results are evaluated on the hard subsets (described in sec \ref{sec:appendix_sub}) from the MVTecAD dataset. 

 \begin{table}[t]
\centering
\caption{AUROC and PRO results on the MVTecAD dataset according to the hyperparameter $\lambda$.}
\label{tab:labmda}
\begin{tabular}{c|ccc}
\hline
$\lambda$ & 1 & 5 & 10 \\
\hline
\hline
Image AUROC & 0.983 & 0.984 & 0.985 \\
\hline
Pixel AUROC & 0.984 & 0.985 & 0.985 \\
\hline
PRO & 0.945 & 0.947 & 0.950 \\
\hline
\end{tabular}
\end{table}

\section{BG-SPP Loss Analysis}
We can transform the first part of our BG-SPP loss as follows:

\begin{align}
    \sum_{i=1}^{N}|{\rm min}((&{\rm log}p_i-b_n),0)| \nonumber \\
    &=\sum_{i=1}^{N}|{\rm max}((-{\rm log}p_i+b_n),0)| \nonumber \\
    &=\sum_{i=1}^{N}|-b_n||{\rm max}(({\rm log}p_i/b_n-1),0)| \nonumber \\
     &=\sum_{i=1}^{N}|b_n|{\rm max}(({\rm log}p_i/b_n-1),0) \nonumber \\
\end{align}
Note that as $b_n$ is negative, we can't transform $|{\rm max}((-{\rm log}p_i+b_n),0)|$ to $|b_n||{\rm max}((-{\rm log}p_i/b_n+1),0)|$, which is not correct. If we replace ${\rm log}p_i/b_n$ by $z$, the first part of our BG-SPP loss can be rewritten as $\sum_{i=1}^{N}|b_n|{\rm max}((z-1),0)$, which can be seen as a rescaled hinge loss (${\rm max}(1-z,0)$) but with opposite optimization direction. The second part of our BG-SPP loss can also be transformed to $\sum_{i=1}^{N}|b_n-\tau|{\rm max}((-{\rm log}p_i/(b_n-\tau)+1),0)$, which can be seen as a rescaled version of the hinge loss.

\section{RandAugment-based Pseudo Anomaly Generation}
\label{sec:appendix_rcp}
\begin{figure}[ht]
    \centering
    \includegraphics[width=1.0\linewidth]{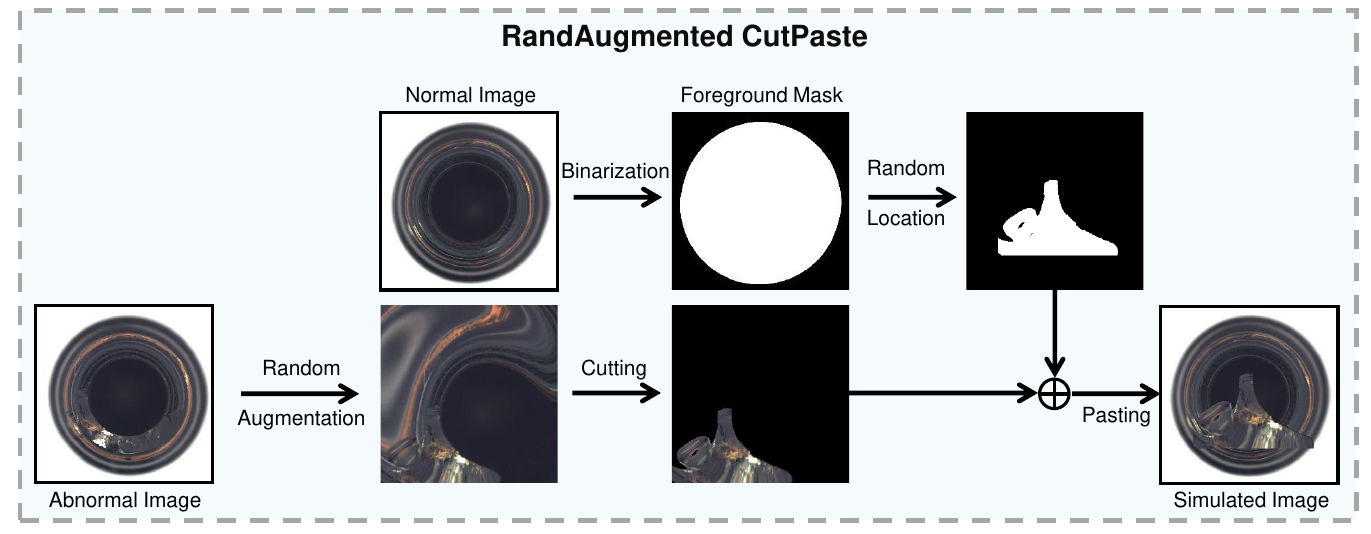} \\
    \caption{Process to generate abnormal samples.}
    \label{fig:data_strategy}
\end{figure}

\begin{figure}[ht]
    \centering
    \includegraphics[width=1.0\linewidth]{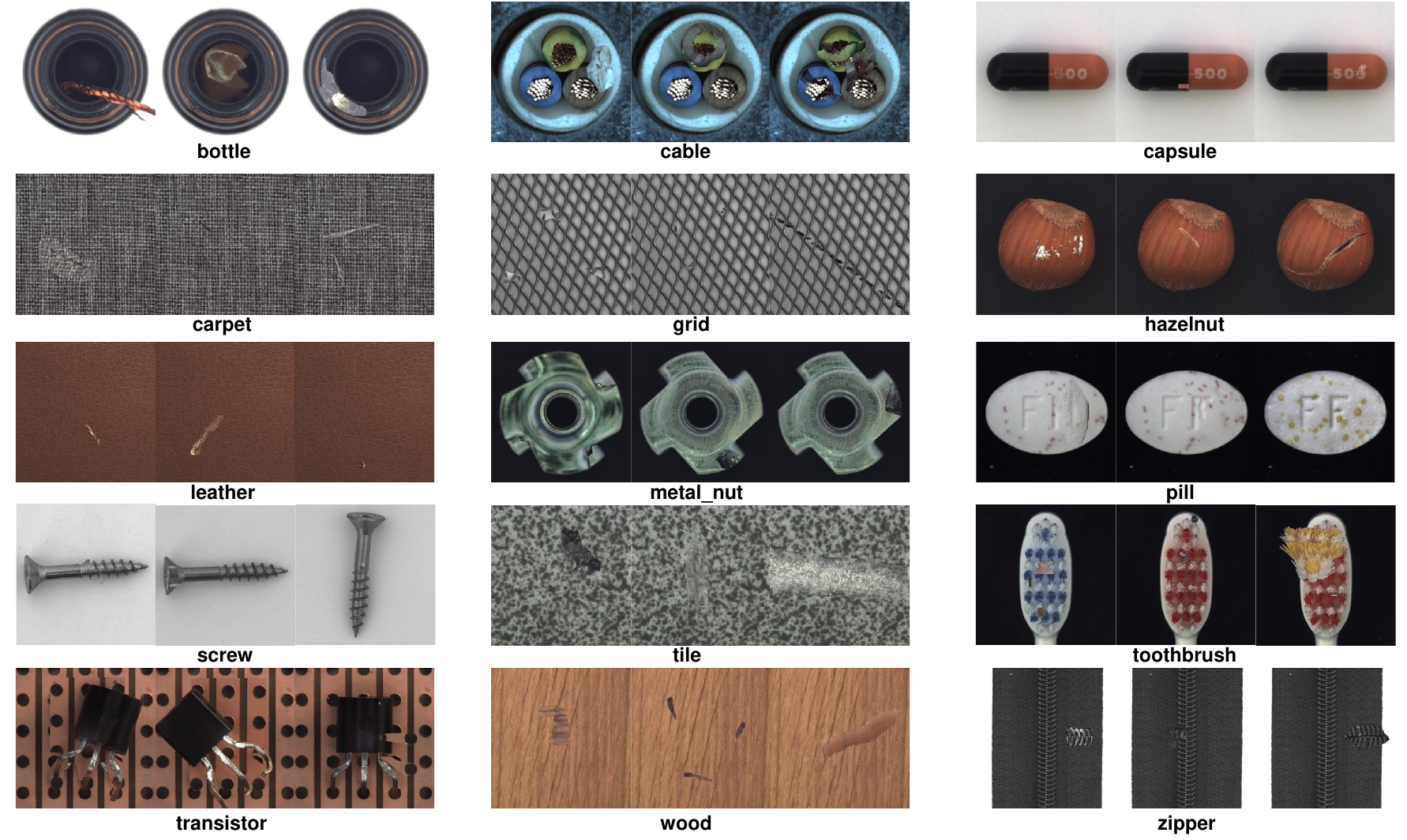} \\
    \caption{Generated abnormal samples by RPAG. All product categories of the MVTecAD dataset are shown in the Figure.}
    \label{fig:aug_images}
\end{figure}

The whole procedure of RandAugment-based Pseudo Anomaly Generation (RPAG) is illustrated in Figure \ref{fig:data_strategy}. More generated abnormal samples by RPAG are shown in Figure \ref{fig:aug_images}.

The advantage of RPAG is that learning from generated samples to recognize irregularities can generalize well to unseen anomalies. The limitation is that RPAG is still not a perfect imitation of real anomalies.

\textbf{Effect of RandAugment-based Pseudo Anomaly Generation.} To show the effectiveness of RandAugment-based Pseudo Anomaly Generation, we show experimental results with or without RPAG in Table \ref{tab:ablation_data_strategy}. It can be found that our method can effectively improve the detection performance even if only five anomalies per category are used. Under the setting of five anomalies, the results can be improved by 0.3\%, 0.3\%, and 1.0\% for image-level AUROC, pixel-level AUROC, and PRO, respectively. Under the setting of ten anomalies, the results can be improved by 0.5\%, 0.3\%, and 0.7\% for image-level AUROC, pixel-level AUROC, and PRO, respectively.

\textbf{Comparison with Other Pseudo Anomaly Generation Strategies.} In DRAEM \cite{DRAEM} and NSA \cite{NSA}, the authors also attempt
to utilize synthetic anomalies, we compare our RPAG strategy
with their strategies and show results in Table \ref{tab:comparison_data_strategy}. The strategy in \cite{DRAEM} is by using texture samples to simulate anomaly regions and by using Perlin noise to capture a variety of anomaly shapes. This strategy may generate invalid anomalies (i.e., anomalies appear in the background) as it will generate anomaly areas in the whole image, while our strategy can ensure to generate more valid anomalies by region limitation. NSA \cite{NSA} integrates Poisson image editing to seamlessly blend scaled normal patches of various sizes from other normal images. By contrast, the anomalies simulated by our strategy are more realistic than those in \cite{NSA} as it’s based on a few real anomalies. Compared with these anomaly generation strategies, the RPAG is more suitable for our method as it can exploit a few known anomalies more sufficiently.

\begin{table}[ht]
\centering
    \caption{The ablation study on the MVTecAD dataset to verify the effectiveness of RandAugment-based Pseudo Anomaly Generation and Asymmetric Weighting. n: n existing anomalies used in training.}
\label{tab:ablation_data_strategy}
\resizebox{\linewidth}{!}{\begin{tabular}{cccc|c|c|c}
\hline
5 & 10 & RPAG & AW & Image AUROC & Pixel AUROC & PRO \\
\hline
\hline
&  &  &  &  0.968 & 0.979 & 0.946\\
\checkmark &  &  &  &  0.982 & 0.985 & 0.959\\
\checkmark & & \checkmark & &  0.985 & 0.988 & 0.969\\
 & \checkmark & & &  0.983 & 0.991 & 0.969\\
 & \checkmark & \checkmark & &  0.988 & 0.992 & 0.973\\
 &  \checkmark & \checkmark & \checkmark & \textbf{0.993} & \textbf{0.992} & \textbf{0.976}\\
\hline
\end{tabular}}
\end{table}

\begin{table}[ht]
\small
\centering
    \caption{Comparison of our RPAG strategy with the pseudo anomaly generation strategies in DRAEM \cite{DRAEM} and NSA \cite{NSA}.}
\label{tab:comparison_data_strategy}
\begin{tabular}{c|c|c|c}
\hline
Strategy & Image AUROC & Pixel AUROC & PRO \\
\hline
\hline
DRAEM \cite{DRAEM} &  0.970 & 0.979 & 0.946\\
NSA \cite{NSA} &  0.977 & 0.979 & 0.946\\
RPAG &  0.988 & 0.992 & 0.973\\
\hline
\end{tabular}
\end{table}

\section{Asymmetric Weighting}
We further propose Asymmetric Weighting (AW) for the objective function to focus on hard normal features and abnormal features to mitigate the rarity problem.

\textbf{Weighting for Hard Normal Features.} For easy normal features, the weights are assigned as $1$. For hard normal features, higher weights should be assigned. Let $\alpha_n$ and $\gamma_n$ are the normal focusing parameters, we propose Truncated Focal Weighting as follows:
\begin{equation}
\label{eq:tfw}
 Fw_{n_i} = \begin{cases}
    1.0, & \text{if}  \quad {\rm log}p_i > {\rm log}p_n; \\
    -\alpha_n(1-p_i)^{\gamma_n} {\rm log}p_i, & \text{if} \quad {\rm log}p_i \leq {\rm log}p_n.
    \end{cases}
\end{equation}

\textbf{Weighting for Abnormal Features.} Abnormal features with larger log-likelihood can be regarded as hard positives. We propose Reversed Focal Weighting to assign higher weights for abnormal features and much higher weights for hard positives. However, the weighting factors may be less than $1$ for abnormal features with smaller log-likelihoods in Reversed Focal Weighting. Therefore, we introduce a truncation term, $1$ will be assigned as weights for easy abnormal features. Let $\alpha_a$ and $\gamma_a$ are the normal focusing parameters, the weighting formula is defined as follows:

\begin{equation}
\label{eq:rfw}
 Fw_{a_j} = \begin{cases}
    -\alpha_a(1+p_j)^{\gamma_a} \frac{1}{{\rm log}p_j}, & \text{if}  \quad {\rm log}p_j > {\rm log}p_a; \\
    1.0, & \text{if} \quad {\rm log}p_j \leq {\rm log}p_a.
    \end{cases}
\end{equation}

\textbf{Hyperparameter Settings.} In E.q.(\ref{eq:tfw}), we set ${\rm log}p_n=-2$ (features with log-likelihoods larger than $-2$ can be regarded as easy normal features empirically), and $\alpha_n=15,\gamma_n=1$ to make $Fw_{n_i}$ more smooth at ${\rm log}p_i=-2$. In E.q.(\ref{eq:rfw}), we set ${\rm log}p_a=-20$ (features with log-likelihoods less than $-20$ can be regarded as easy abnormal features empirically), and $\alpha_a=0.53, \gamma_a=2$ to make $Fw_{a_j}$ more smooth at ${\rm log}p_j=-20$.  

\textbf{Detailed Weighted Learning Objective.} The detailed weighted learning objective is formulated as follows:
\begin{align}
    \mathcal{L} &= \mathcal{L}_{ml} \odot \mathcal{FW}_1 + \lambda \mathcal{L}_{bg-spp} \odot \mathcal{FW}_2 \nonumber \\
    & = \mathbb{E}_{x \in \mathcal{X}^n}\Big[Fw_{n_i}\cdot(\frac{1}{2}\varphi_\theta(x)^T\varphi_\theta(x)\nonumber \\ &-\sum\nolimits_{l=1}^{L}{\rm log}|{\rm det}J_{\varphi_l}(y_{l-1})|+\frac{d}{2}{\rm log}(2\pi))\Big] \nonumber \\
    & + \sum_{i=1}^{|\mathcal{X}^n|}Fw_{n_i}\cdot|{\rm min}(({\rm log}p_i-b_n),0)| \nonumber \\ & + \sum_{j=1}^{|\mathcal{X}^a|} Fw_{a_j}\cdot|{\rm max}(({\rm log}p_j-b_n+\tau), 0)|
\end{align}

\textbf{Effect of Asymmetric Weighting.} As shown in Table \ref{tab:ablation_data_strategy}, the detection and localization performance can be further improved by Asymmetric Weighting. The experimental results show that RandAugment-based Pseudo Anomaly Generation and Asymmetric Weighting can mitigate the rarity problem effectively.

\section{More Results under the One-Class Setting}
In Table \ref{tab:unseen_setting_append}, we show more results under the one-class setting. All the other results are from \cite{DRA}. However, in \cite{DRA}, only image-level AUROCs are reported, and the results of some categories shown in Table \ref{tab:unseen_setting_append} are missing.

\textbf{Comparison to Unsupervised Baseline.} Our BGAD can outperform the baseline NFAD across all the datasets, especially on the object categories with more complex normal patterns (\emph{e.g.}, Capsule, Screw, Transistor). This shows our model’s better generalizability to unseen anomalies.

\begin{table}
\caption{AUC results under the one-class setting, where models are trained with only one anomaly class and tested to detect other anomaly classes. $\cdot/\cdot$ means image-level and pixel-level AUROCs.}
\label{tab:unseen_setting_append}
\resizebox{1.0\linewidth}{!}{
\begin{tabular}{cc|c|cccccc}
\hline
 \multirow{2}*{Dataset} & \multirow{2}*{Known Class} & Baseline  & \multicolumn{6}{c}{Ten Training Anomaly Samples} \\
 \cline{3-9}
 & & NFAD & DevNet  & FLOS  & SAOE  & MLEP  & DRA  & BGAD (Ours) \\
\hline
 \multirow{7}*{\rotatebox{90}{AITEX}} & Broken\_end & 0.835/\textbf{0.828} & 0.658/- & 0.585/- & 0.712/- & 0.732/- & 0.693/- & \textbf{0.856}/0.824\\
 & Broken\_pick & \textbf{0.960}/0.982 & 0.585/- & 0.548/-  & 0.629/- & 0.555/- & 0.760/- & 0.948/\textbf{0.983}\\
 & Cut\_selvage & 0.834/0.830 & 0.709/- & 0.745/- & 0.770/-  & 0.682/- & 0.777/- & \textbf{0.865}/\textbf{0.844}\\
 & Fuzzyball & 0.815/\textbf{0.827}  & 0.734/- & 0.550/- & 0.842/- & 0.677/- & 0.701/- & \textbf{0.823}/0.825\\
 & Nep & 0.834/\textbf{0.828} & 0.810/- & 0.746/- & 0.771/- & 0.740/- & 0.750/- & \textbf{0.838}/0.825\\
 & Weft\_crack & 0.827/\textbf{0.692} & 0.599/- & 0.636/- & 0.618/- & 0.370/- & 0.717/- & \textbf{0.841}/0.685\\
 \cline{2-9}
 & \textbf{Mean} & 0.851/\textbf{0.831} & 0.683/- & 0.635/-  & 0.724/- & 0.626/- & 0.733/- &  \textbf{0.862}/\textbf{0.831}\\
 \hline
 \multirow{3}*{\rotatebox{90}{ELPV}} & Mono & 0.860/- & 0.599/- & 0.629/- & 0.569/- & 0.756/- & 0.731/- & \textbf{0.884}/-\\
 & Poly & 0.870/-  & 0.804/- & 0.662/-  & 0.796/- & 0.734/- & 0.800/- & \textbf{0.880}/-\\
 \cline{2-9}
 & \textbf{Mean} & 0.865/-  & 0.702/- & 0.646/-  & 0.683/- & 0.745/- & 0.766/- &  \textbf{0.882}/-\\
 \hline
 \multirow{4}*{\rotatebox{90}{Bottle}} & Broken\_large & \textbf{1.000}/0.989 & -/- & -/- & -/- & -/- & -/- & \textbf{1.000}/\textbf{0.991}\\
 & Broken\_small & \textbf{1.000}/0.987  & -/- & -/-  & -/- & -/- & -/- & \textbf{1.000}/\textbf{0.988}\\
 & Contamination & \textbf{1.000}/0.994  & -/- & -/-  & -/- & -/- & -/- & \textbf{1.000}/\textbf{0.996}\\
 \cline{2-9}
 & \textbf{Mean} & \textbf{1.000}/0.990 & -/- & -/-  & -/- & -/- & -/- &  \textbf{1.000}/\textbf{0.992}\\
 \hline
 \multirow{6}*{\rotatebox{90}{Capsule}} & Crack & 0.934/0.990  & -/- & -/- & -/- & -/- & -/- & \textbf{0.984}/\textbf{0.991}\\
 & Imprint & 0.955/0.992  & -/- & -/-  & -/- & -/- & -/- & \textbf{0.990}/\textbf{0.993}\\
 & Poke & 0.936/0.989 & -/- & -/-  & -/- & -/- & -/- & \textbf{0.984}/\textbf{0.991}\\
 & Scratch & 0.951/0.989 & -/- & -/-  & -/- & -/- & -/- & \textbf{0.994}/\textbf{0.990}\\
 & Squeeze & 0.928/0.990 & -/- & -/-  & -/- & -/- & -/- & \textbf{0.988}/\textbf{0.991}\\
 \cline{2-9}
 & \textbf{Mean} & 0.941/0.990  & -/- & -/-  & -/- & -/- & -/- &  \textbf{0.988}/\textbf{0.991}\\
 \hline
 \multirow{6}*{\rotatebox{90}{Grid}} & Bent & \textbf{0.990}/0.994  & -/- & -/- & -/- & -/- & -/- & 0.989/\textbf{0.995}\\
 & Broken & 0.982/0.993  & -/- & -/-  & -/- & -/- & -/- & \textbf{0.994}/\textbf{0.994}\\
 & Glue & 0.990/0.993 & -/- & -/-  & -/- & -/- & -/- & \textbf{1.000}/\textbf{0.994}\\
 & Metal & 0.982/0.993 & -/- & -/-  & -/- & -/- & -/- & \textbf{0.992}/\textbf{0.994}\\
 & Thread & 0.982/0.995 & -/- & -/-  & -/- & -/- & -/- & \textbf{0.990}/\textbf{0.996}\\
 \cline{2-9}
 & \textbf{Mean} & 0.985/0.994 & -/- & -/-  & -/- & -/- & -/- &  \textbf{0.993}/\textbf{0.995}\\
 \hline
 \multirow{5}*{\rotatebox{90}{Hazelnut}} & Crack & \textbf{1.000}/0.996 & -/- & -/- & -/- & -/- & -/- & \textbf{1.000}/\textbf{0.997}\\
 & Cut & \textbf{1.000}/0.984 & -/- & -/-  & -/- & -/- & -/- & \textbf{1.000}/\textbf{0.985}\\
 & Hole & 0.997/0.981 & -/- & -/-  & -/- & -/- & -/- & \textbf{1.000}/\textbf{0.985}\\
  & Print & 0.997/0.981 & -/- & -/-  & -/- & -/- & -/- & \textbf{0.999}/\textbf{0.984}\\
 \cline{2-9}
 & \textbf{Mean} & 0.998/0.985 & -/- & -/-  & -/- & -/- & -/- &  \textbf{1.000}/\textbf{0.988}\\
 \hline
 \multirow{6}*{\rotatebox{90}{Screw}} & Manipulated & 0.887/0.991 & -/- & -/- & -/- & -/- & -/- & \textbf{0.957}/\textbf{0.993}\\
 & Scratch\_head & 0.874/0.988 & -/- & -/-  & -/- & -/- & -/- & \textbf{0.927}/\textbf{0.990}\\
 & Scratch\_neck & 0.861/0.987 & -/- & -/-  & -/- & -/- & -/- & \textbf{0.944}/\textbf{0.989}\\
  & Thread\_side & 0.927/0.992 & -/- & -/-  & -/- & -/- & -/- & \textbf{0.965}/\textbf{0.994}\\
  & Thread\_top & 0.878/0.988 & -/- & -/-  & -/- & -/- & -/- & \textbf{0.943}/\textbf{0.989}\\
 \cline{2-9}
 & \textbf{Mean} & 0.885/0.989 & -/- & -/-  & -/- & -/- & -/- &  \textbf{0.947}/\textbf{0.991}\\
 \hline
 \multirow{6}*{\rotatebox{90}{Tile}} & Crack & 0.997/\textbf{0.979} & -/- & -/- & -/- & -/- & -/- & \textbf{1.000}/0.978\\
 & Glue\_strip & 0.997/0.967 & -/- & -/-  & -/- & -/- & -/- & \textbf{1.000}/\textbf{0.979}\\
 & Gray\_stroke & 0.999/0.963 & -/- & -/-  & -/- & -/- & -/- & \textbf{1.000}/\textbf{0.969}\\
  & Oil & 0.997/0.959 & -/- & -/-  & -/- & -/- & -/- & \textbf{1.000}/\textbf{0.967}\\
   & Rough & 0.998/0.977 & -/- & -/-  & -/- & -/- & -/- & \textbf{1.000}/\textbf{0.987}\\
 \cline{2-9}
 & \textbf{Mean} & 0.998/0.969 & -/- & -/-  & -/- & -/- & -/- &  \textbf{1.000}/\textbf{0.976}\\
 \hline
 \multirow{5}*{\rotatebox{90}{Transistor}} & Bent\_lead & 0.979/\textbf{0.925} & -/- & -/- & -/- & -/- & -/- & \textbf{0.988}/0.921\\
 & Cut\_lead & 0.982/0.927 & -/- & -/-  & -/- & -/- & -/- & \textbf{1.000}/\textbf{0.935}\\
 & Damaged & 0.985/\textbf{0.921} & -/- & -/-  & -/- & -/- & -/- & \textbf{0.989}/0.919\\
 & Misplaced & 0.991/0.945 & -/- & -/-  & -/- & -/- & -/- & \textbf{1.000}/\textbf{0.994}\\
 \cline{2-9}
 & \textbf{Mean} & 0.984/0.929 & -/- & -/-  & -/- & -/- & -/- &  \textbf{0.994}/\textbf{0.942}\\
 \hline
 \multirow{6}*{\rotatebox{90}{Wood}} & Color & \textbf{0.995}/0.968 & -/- & -/- & -/- & -/- & -/- & 0.994/\textbf{0.974}\\
 & Combined & \textbf{0.994}/0.970 & -/- & -/-  & -/- & -/- & -/- & \textbf{0.994}/\textbf{0.977}\\
 & Hole & 0.994/0.968 & -/- & -/-  & -/- & -/- & -/- & \textbf{0.999}/\textbf{0.973}\\
  & Liquid & \textbf{0.994}/0.967 & -/- & -/-  & -/- & -/- & -/- & 0.992/\textbf{0.970}\\
   & Scratch & \textbf{1.000}/0.986 & -/- & -/-  & -/- & -/- & -/- & 0.999/\textbf{0.988}\\
 \cline{2-9}
 & \textbf{Mean} & \textbf{0.995}/0.972 & -/- & -/-  & -/- & -/- & -/- &  \textbf{0.995}/\textbf{0.976}\\
\hline
\hline
\end{tabular}}
\end{table}

\section{Effect of Semi-Push-Pull Mechanism}
In Table \ref{tab:semi_push_pull_append}, we show more comparison results between BGAD$^{\dag}$ and BGAD. As shown in Table \ref{tab:semi_push_pull_append}, the BGAD$^{\dag}$ performs less effective than the BGAD, and even worse than the baseline NFAD. The comparison between BGAD and BGAD$^{\dag}$ shows that the semi-push-pull mechanism in BGAD is critical for mitigating the bias issue.

\begin{table}
\caption{AUROC results under the one-class setting. $\cdot/\cdot$ means image-level and pixel-level AUROCs.}
\label{tab:semi_push_pull_append}
\resizebox{1.0\linewidth}{!}{
\begin{tabular}{c|ccccc}
\hline
 \multirow{2}*{Method} & \multicolumn{5}{c}{Category} \\
 \cline{2-6}
 & Bottle & hazelnut & Grid  & Tile & Wood\\
\hline
 NFAD (baseline) & \textbf{1.000}/0.990 & 0.998/0.986 & 0.985/0.994 & 0.998/0.969 & \textbf{0.995}/0.972 \\
  BGAD$^{\dag}$ & \textbf{1.000}/0.983 & \textbf{1.000}/0.978 & 0.987/0.988  & 0.999/0.967 & 0.994/0.969 \\
  BGAD (Ours) & \textbf{1.000}/\textbf{0.992} & \textbf{1.000}/\textbf{0.988} & \textbf{0.993}/\textbf{0.995}  & \textbf{1.000}/\textbf{0.976} & \textbf{0.995}/\textbf{0.976} \\
\hline
\hline
\end{tabular}}
\end{table}

\section{Hyper-parameter Sensitivity}
 The main tunable hyperparameters of our model are the normal boundary (controlled by $\beta$) and the abnormal boundary (controlled by $\tau$). As shown in Table \ref{tab:hyperparameter_a}, we evaluate different combinations of $\beta$ (1\%, 5\%, 10\%) and $\tau$ (0.1, 0.2, 0.3).  From Table \ref{tab:hyperparameter_a}, we can draw the following main conclusions: 1) $\beta$ has a more significant effect on performance compared with $\tau$, and pixel-level AUROC is insensitive to the hyperparameters. 2) Our model is not very sensitive to the margin $\tau$, which means our model can achieve superior results as long as a certain margin is formed between normal and abnormal. 

\begin{table}[t]
     \caption{AUROC and PRO results on the MVTecAD dataset according to hyperparameter $\beta$ and $\tau$. $\cdot/\cdot/\cdot$ means mean image-level AUROC, mean pixel-level AUROC and mean PRO.}
\label{tab:hyperparameter_a}
\resizebox{\linewidth}{!} {
\begin{tabular}{c|ccc}
\hline
\diagbox{$\beta$}{$\tau$} & 0.1 & 0.2 & 0.3\\
\hline
\hline
1\% & \textbf{0.9936}/0.9920/0.9749 & 0.9935/0.9920/0.9752 & 0.9930/0.9918/0.9748 \\
5\% & 0.9916/0.9922/0.9759 & 0.9918/\textbf{0.9923}/0.9763 & 0.9922/0.9921/0.9762 \\
10\% & 0.9915/0.9922/0.9759 & 0.9920/0.9922/\textbf{0.9764} & 0.9925/0.9922/0.9762\\
\hline
\end{tabular}}
 \end{table}

\section{Learning Efficiency}

\begin{figure}[ht]
    \centering
    \includegraphics[width=0.8\linewidth]{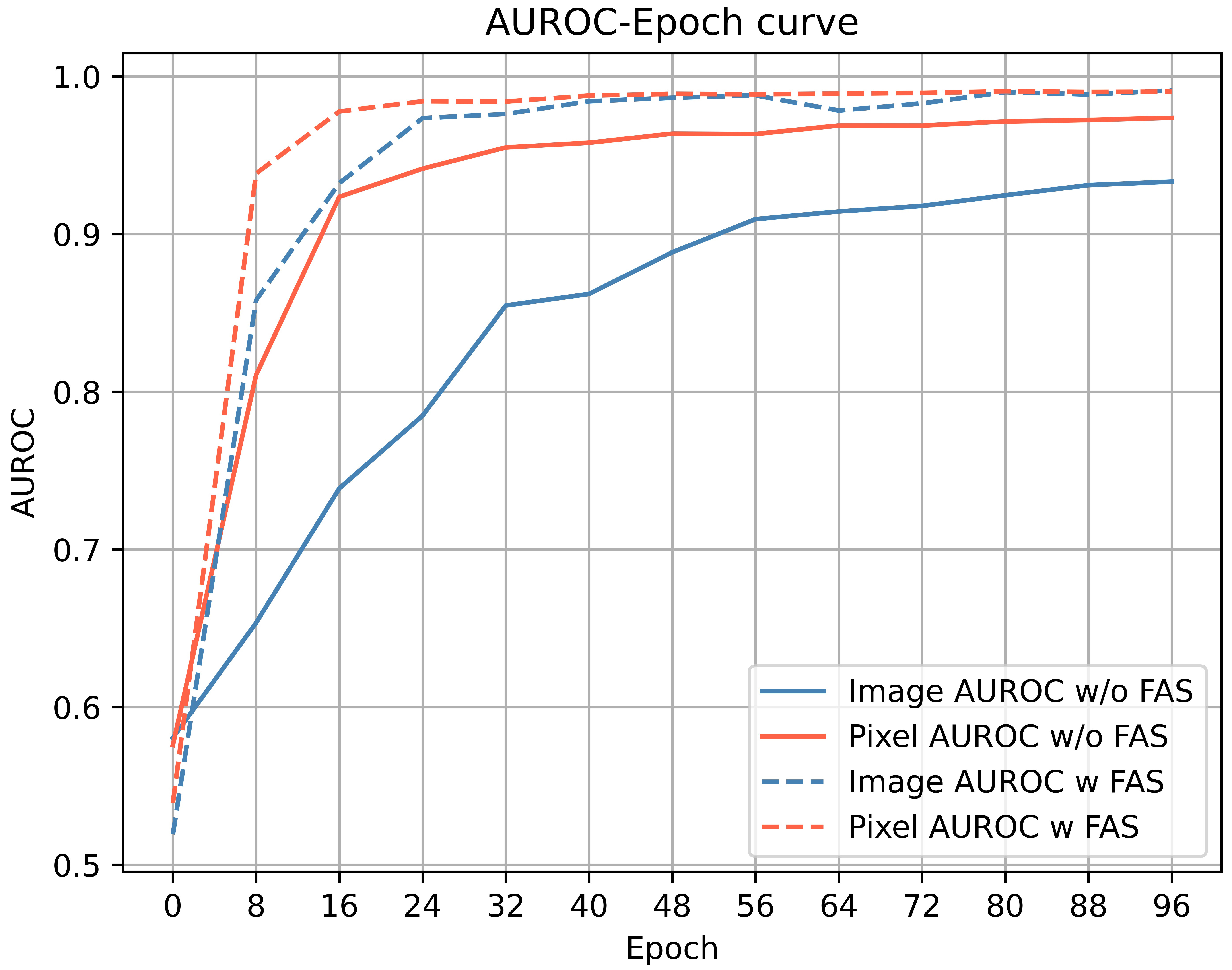}
    \caption{AUROC vs epoch curve of cable category on the MVTecAD dataset.}
    \label{fig:learning_efficiency}
\end{figure}

In addition to the improvement of detection results, our method can also achieve significant improvement in learning efficiency. To illustrate the learning efficiency, we show AUROC vs epoch curve in Figure \ref{fig:learning_efficiency}, specifically, the pixel-level AUROC with a few abnormal samples (FAS) converges rapidly compared to its counterparts. The AUROC can increase a large margin generally only a meta epoch (8 epochs) after adding BG-SPP loss for optimization. 

\section{Details of Hard and Unseen Subset Selection}
\label{sec:appendix_sub}
In order to thoroughly verify the effectiveness of our method, we further construct two more difficult subsets from the MVTecAD dataset. The first subset is constructed to evaluate the detection performance, thus we select the subset based on the image-level AUROC. Specifically, we select the first subset based on the misclassification at the image level, \emph{i.e.} anomaly categories are selected if several samples of these categories are detected as normal. The second subset is constructed to evaluate the localization performance, thus we select the subset based on the pixel-level AUROC. Specifically, anomaly categories are selected if their pixel-level AUROCs are the lowest among all anomaly categories. The constructed subsets are shown in Table \ref{tab:subsets} (Note: As there is only one anomaly class in the toothbrush category, for simplicity, we set the easy and hard subset as the same). In order to verify the generalization capability of our model, we only use the easy subsets as the training set and validate results on the hard subsets. Thus, the hard subsets are utilized as the unseen subsets for generalization capability evaluation.  

\begin{table*}[ht]
\caption{Two hard subsets constructed from the MVTecAD dataset.}
\label{tab:subsets}
\resizebox{1.0\linewidth}{!}{
\begin{tabular}{cccc|cc}
\hline
& &\multicolumn{2}{c|}{First Subset} & \multicolumn{2}{c}{Second Subset} \\
 & Category & Easy Anomaly Categories & Hard Anomaly Categories & Easy Anomaly Categories & Hard Anomaly Categories \\
\hline
 \multirow{5}*{\rotatebox{90}{Textures}}& Carpet &  color, cut, hole, metal\_contamination & thread  &  color, cut, hole, metal\_contamination& thread\\
  & Grid & broken, metal\_contamination, thread & glue, bent & bent, broken, glue,  metal\_contamination & thread\\
  & Leather & color, fold, glue, poke & cut & color, cut, glue, poke & fold\\
  & Tile & crack, glue\_strip, gray\_stroke, oil & rough & crack, glue\_strip, gray\_stroke, oil & rough \\
  & Wood & color, combined, hole, liquid & scratch & color, combined, hole, liquid & scratch\\
\hline
 \multirow{10}*{\rotatebox{90}{Objects}} & Bottle & broken\_large, broken\_small &  contamination & broken\_large, broken\_small & contamination\\
 & Cable & bent\_wire, combined, cut\_inner\_insulation,cut\_outer\_insulation,missing\_cable & cable\_swap, missing\_wire, poke\_insulation & bent\_wire, cable\_swap, combined, cut\_inner\_insulation, missing\_cable, missing\_wire, poke\_insulation & cut\_outer\_insulation\\
 & Capsule & crack, squeeze & faulty\_imprint, poke, scratch & crack, faulty\_imprint, poke, scratch & squeeze\\
 & Hazelnut & crack, hole, print & cut & cut, hole, print & crack\\
 & Metal nut & bent, color, flip & scratch & bent, color, scratch & flip\\
 & Pill & color, combined, contamination, faulty\_imprint, pill\_type & crack, scratch & color, combined, contamination, crack, faulty\_imprint, scratch & pill\_type\\
 & Screw & scratch\_head, scratch\_neck, thread\_top & manipulated\_front, thread\_side & manipulated\_front, scratch\_head, scratch\_neck, thread\_top & thread\_side\\
 & Toothbrush & defective & defective & defective & defective\\
 & Transistor & bent\_lead, cut\_lead, misplaced & damaged\_case & bent\_lead, damaged\_case, misplaced & cut\_lead\\
 & Zipper & broken\_teeth, combined, fabric\_border, rough, split\_teeth & fabric\_interior, squeezed\_teeth & broken\_teeth, combined, fabric\_border, fabric\_interior, split\_teeth, squeezed\_teeth & rough\\
\hline
\end{tabular}}
\end{table*}
\end{document}